\documentclass{bmvc2k}

\title{Incremental Multi-Scene Modeling via Continual Neural Graphics Primitives}

\addauthor{Prajwal Singh}{singh_prajwal@iitgn.ac.in}{1}
\addauthor{Ashish Tiwari}{ashish.tiwari@iitgn.ac.in}{1}
\addauthor{Gautam Vashishtha}{gautam.pv@iitgn.ac.in}{1}
\addauthor{Shanmuganathan Raman}{shanmuga@iitgn.ac.in}{1}

\addinstitution{
 CVIG Lab\\
 IIT Gandhinagar\\
 Gujarat, India
}

\runninghead{\textit{SINGH et.al.}}{\textit{Incremental Multi-Scene Modeling via C-NGP}}

\usepackage{graphicx}	
\usepackage{amsmath}	
\usepackage{amssymb}	
\usepackage{booktabs}
\usepackage{times}
\usepackage{microtype}
\usepackage{epsfig}
\usepackage{caption}
\usepackage{float}
\usepackage{placeins}
\usepackage{color, colortbl}
\usepackage{stfloats}
\usepackage{enumitem}
\usepackage{tabularx}
\usepackage{xstring}
\usepackage{multirow}
\usepackage{xspace}
\usepackage{url}
\usepackage{subcaption}
\usepackage[hang,flushmargin]{footmisc}

\usepackage{soul}
\usepackage{changepage,threeparttable} %

\usepackage[ruled,vlined]{algorithm2e}

% ...

\newcommand{\R}[1]{{%
    \textbf{%
        \ifstrequal{#1}{1}{\textcolor{red}{R#1}}{%
        \ifstrequal{#1}{2}{\textcolor{blue}{R#1}}{%
        \ifstrequal{#1}{3}{\textcolor{magenta}{R#1}}{%
        \ifstrequal{#1}{4}{\textcolor{teal}{R#1}}{%
                           \textcolor{cyan}{R#1}%
        }}}}%
    }%
}}

\SetCommentSty{mycommfont}

\SetKwInput{KwInput}{Require}                
\SetKwInput{KwOutput}{Output} 

\makeatletter
\newcommand{\removelatexerror}{\let\@latex@error\@gobble}
\makeatother

\begin{document}

\maketitle

\begin{abstract}
Neural radiance fields (NeRF) have revolutionized photorealistic rendering of novel views for 3D scenes. Despite their growing popularity and efficiency as 3D resources, NeRFs face scalability challenges due to the need for separate models per scene and the cumulative increase in training time for multiple scenes. The potential for incrementally encoding multiple 3D scenes into a single NeRF model remains largely unexplored. To address this, we introduce Continual-Neural Graphics Primitives (C-NGP), a novel continual learning framework that integrates multiple scenes incrementally into a single neural radiance field. Using a generative replay approach, C-NGP adapts to new scenes without requiring access to old data. We demonstrate that C-NGP can accommodate multiple scenes without increasing the parameter count, producing high-quality novel-view renderings on synthetic and real datasets. Notably, C-NGP models all $8$ scenes from the Real-LLFF dataset together, with only a $2.2\%$ drop in PSNR compared to vanilla NeRF, which models each scene independently. Further, C-NGP allows multiple style edits in the same network. The code implementation and dynamic visualizations can be accessed from here \url{https://prajwalsingh.github.io/C-NGP/}. 
\end{abstract}


\begin{figure}
\centering
\includegraphics[width=0.9\linewidth]{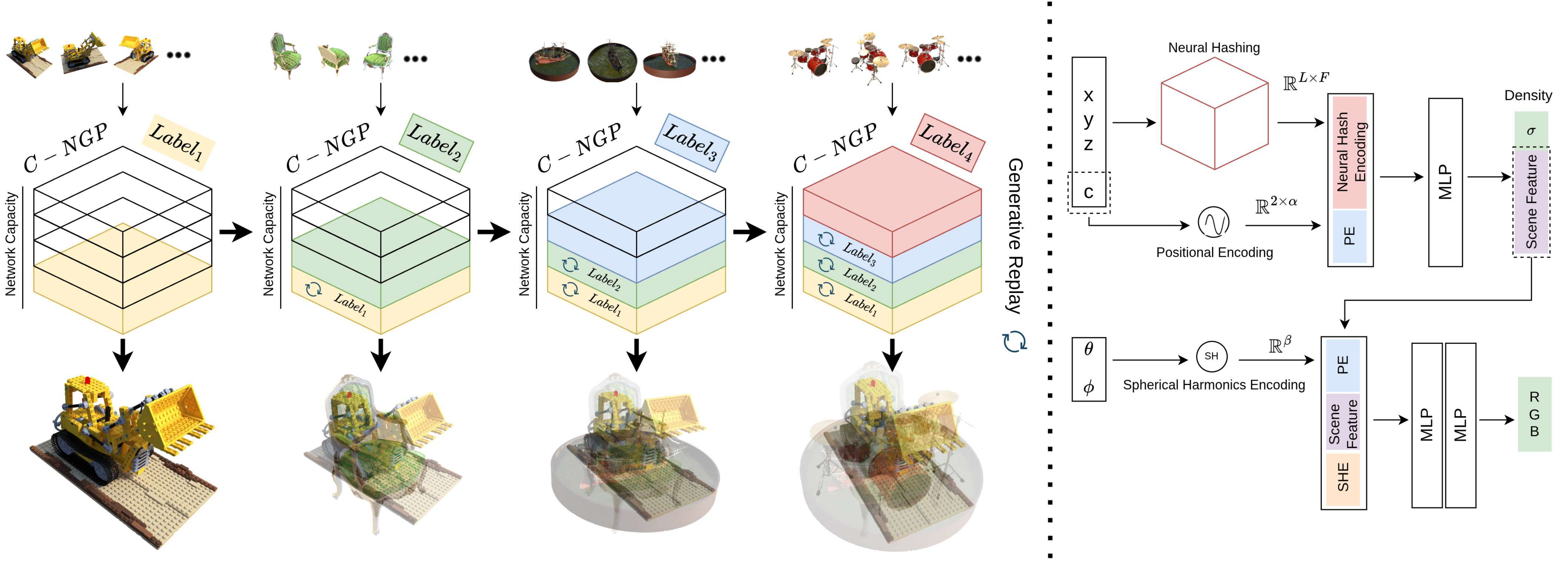}
\vspace{0.5em}
\caption{(Left) The proposed continual neural graphics primitives (C-NGP) framework to model multiple scenes. Every new scene is assigned a unique pseudo-label and trained sequentially while preserving the previously learned information via generative replay. Any specific scene can be rendered by conditioning C-NGP over the associated pseudo-label. (Right) Our modification to the Instant-NGP architecture \cite{muller2022instant} enables multi-scene modeling using pseudo labels.
}
\label{fig:framework}
\vspace{-0.55cm}
\end{figure}

\section{Introduction}
\label{sec:introduction}

Neural Radiance Fields (NeRF) synthesize photorealistic novel views by modeling 3D scenes using Multi-Layer Perceptrons (MLPs) and differentiable volumetric rendering. While powerful, NeRF and its variants typically require per-scene optimization, leading to high training and storage costs. Recent methods address speed \cite{wang2021neus, fridovich2022plenoxels, muller2022instant}, inference efficiency \cite{chen2023mobilenerf, wan2023learning}, compression \cite{gordon2023quantizing, chen2022tensorf, tang2022compressible, shin2024binary, fang2023one}, scene editing \cite{wang2023rip, bao2023sine}, and generalization to unseen scenes \cite{he2023cp, yu2021pixelnerf, wang2021ibrnet, zhang2022nerfusion, yang2023contranerf, chen2023gm}. However, learning multiple distinct scenes within a single NeRF model remains underexplored, with SCARF \cite{wang2024scarf} being a recent attempt.

Training separate NeRFs for each scene is impractical for large-scale 3D collections due to resource demands. We propose a \textit{continual learning} approach that incrementally encodes multiple scenes within a single NeRF while maintaining a fixed parameter count. Our framework learns new scenes sequentially without forgetting previous ones and without having to store past training data. Effective multi-scene modeling requires: \textit{(a)} photorealistic rendering of all scenes, \textit{(b)} integration of new scenes without access to old data, \textit{(c)} prevention of catastrophic forgetting, and \textit{(d)} constant model size. Moreover, continual learning in NeRFs is uniquely difficult. Unlike image classification, where an image seldom has different labels, NeRF requires scene-specific mappings since the same position and view direction across different scenes produce different outputs. 


We introduce C-NGP - \textbf{C}ontinual \textbf{N}eural \textbf{G}raphics \textbf{P}rimitives that leverage fast training and efficient learning of Instant-NGP \cite{muller2022instant} by integrating scene conditioning into its hashing mechanism using pseudo-labels to accommodate multiple scenes into NGP's parameter space. This simple yet effective conditioning enables multi-scene optimization, allowing the model to learn new scenes while preserving high-fidelity rendering of previously encountered scenes. C-NGP uses generative-replay to accommodate new scenes without requiring access to training data from previously learned scenes. Thus, C-NGP is practical for real-world use, allowing multiple 3D scenes to be stored and shared in a single network checkpoint, where the new scenes can be added incrementally without creating separate models. Interestingly, even vanilla NeRF can be adapted to the proposed setting, but with far higher computational overhead (12 days of training time vs. 8 hours for C-NGP) and lower performance.

Our main contributions are: 1) C-NGP, the first fixed-size NeRF to continually encode multiple scenes, achieving only a $2.2\%$ PSNR drop on Real-LLFF \cite{mildenhall2019local} compared to scene-specific vanilla NeRF; 2) training without access to previous data via generative replay; 3) an efficient scene-conditioning mechanism in Instant-NGP’s hash grid; 4) support for multi-view consistent style editing across scenes; 5) competitive performance over continual learning baselines \cite{chung2022meil, cai2023clnerf, po2023instant} and concurrent methods \cite{wang2024scarf} in quality, training efficiency, and adaptability to new scenes.

\section{Related Work}
\label{sec:relatedwork}

\textbf{Single-Scene Neural Radiance Fields.} NeRF encodes scene geometry and appearance but requires long per-scene optimization. Alternatives like Plenoxels \cite{fridovich2022plenoxels} and Instant-NGP \cite{muller2022instant} improve efficiency using spherical harmonics and multi-resolution hash encoding. Other extensions enable faster inference \cite{chen2023mobilenerf, wan2023learning}, scene editing \cite{wang2023rip, bao2023sine}, or rendering from sparse images \cite{lee2024few, yang2023freenerf, wang2023sparsenerf}. Another line of work aims to generalize NeRF to unseen scenes without per-scene training \cite{yu2021pixelnerf, wang2021ibrnet, chen2021mvsnerf, he2023cp, zhang2022nerfusion, yang2023contranerf}. These methods use convolutional feature extractors, 3D cost volumes, or attention modules \cite{park2019deepsdf, barron2022mip} to learn scene priors, enabling inference on new scenes. However, they often require fine-tuning for optimal quality and are not designed for incremental learning, suffering from catastrophic forgetting when adapting to multiple scenes in a shared parameter space. Unlike these approaches, our method focuses on continual learning to incrementally encode distinct scenes without forgetting, targeting compact 3D asset storage rather than generalization. Recently, 3D Gaussian Splatting (3DGS) methods \cite{Kerbl20233D, chen2024mvsplat} have achieved real-time rendering for single scenes but do not support incremental multi-scene learning. However, these approaches continue to focus on single-scene optimization.


\textbf{Continual Learning in NeRF.} Continual learning enables models to learn new tasks sequentially without forgetting prior knowledge \cite{robins1995catastrophic}. Techniques include parameter isolation \cite{mallya2018packnet}, regularization \cite{kirkpatrick2017overcoming, li2017learning}, and replay \cite{shin2017continual, rannen2017encoder}. In NeRF, applying continual learning is challenging due to scene-specific representations, as the same 3D point and view direction yield different densities and colors across scenes. Most existing methods focus on single-scene scenarios, addressing appearance or geometry changes over time \cite{chung2022meil, cai2023clnerf, po2023instant, zhang2023nerf}. For example, MEIL-NeRF \cite{chung2022meil} and CLNeRF \cite{cai2023clnerf} reconstruct scenes from partial scans but are limited to single-scene forgetting. SLAM-based methods \cite{sucar2021imap, chen2023local, deng2024plgslam} assume past data availability for keyframe optimization, unsuitable for multi-scene settings without stored data.

\textbf{Multi-Scene Continual Learning in NeRF.} Encoding multiple distinct scenes in a single NeRF model is largely unexplored. A concurrent work, SCARF \cite{wang2024scarf}, uses a hyper-network to generate scene-specific weights, but this increases the parameter count with each new scene. In contrast, our C-NGP method leverages generative replay \cite{shin2017continual} to learn new scenes without storing past data, maintaining a fixed parameter count. By integrating scene conditioning into Instant-NGP’s \cite{muller2022instant} hashing mechanism, C-NGP achieves high-fidelity rendering and fast convergence, outperforming continual learning baselines \cite{chung2022meil, cai2023clnerf} and competitive performance with SCARF \cite{wang2024scarf} in multi-scene scenarios.
\section{Method}
\label{sec:method}

 We present C-NGP, a framework to incrementally model multiple scenes with the same set of parameters that is traditionally used for modeling a single scene in an NGP network. We begin with a brief overview of learning neural radiance fields and the multi-resolution hash encoding (MHE) technique. Next, we explore the conditional capabilities of Instant-NGP and describe how generative replay \cite{shin2017continual} can be integrated into conditional Instant-NGP. This extension enables incremental learning of new scenes while preserving information from previously encountered scenes. 

\subsection{Background}
\label{sec:background}

\textbf{Neural Radiance Fields.} Mildenhall \emph{et al.} \cite{mildenhall2021nerf} introduced NeRF, a method for synthesizing novel views from a set of input images using a multi-layer perceptron (MLP) that maps a 5D input—3D spatial coordinates $(x, y, z)$ and viewing direction $(\theta, \phi)$—to the density $\sigma$ and color $c$ of a point. Points are sampled along camera rays $\mathbf{r}(t) = \mathbf{o} + t\mathbf{d}$, where $\mathbf{o}$ is the camera origin, $\mathbf{d}$ is the ray direction, and $t$ is the sampled depth. Rendering is performed via volumetric integration \cite{max1995optical}, computing the color of a ray as $\hat{C}(r) = \sum_{i=1}^{N} w_{i} c_{i}$, where the weights $w_i = T_i \alpha_i$ depend on transmittance $T_i$ and opacity $\alpha_i = (1 - e^{-\sigma_i \delta_i})$, with $\delta_i$ being the distance between adjacent samples. The model is trained by minimizing the squared error between predicted and ground-truth pixel colors: $\mathcal{L} = \sum_{r \in \mathcal{R}}|| \hat{C}(r) - C(r)||_{2}^{2}$, where $\mathcal{R}$ denotes the set of rays. Inputs to the MLP are encoded using sinusoidal positional encodings to better capture high-frequency scene details.

\textbf{Multi-Resolution Hash Encoding.} Müller \emph{et al.} \cite{muller2022instant} introduced the concept of multi-resolution hash encoding (MHE) to enhance both the reconstruction accuracy and training efficiency of neural radiance fields (NeRFs) while maintaining minimal computational overhead. Unlike the traditional positional encoding used in NeRF, MHE employs a trainable multi-level 3D grid structure. The position of a sampled point is encoded by interpolating the features $(\mathbb{R}^{F})$ located at the vertices of the grid cell containing the point. These interpolated features are then fed into an MLP network, predicting both the density $(\sigma)$ and the point's color $(c)$.

The grid is organized into $L$ levels, each containing a hash table of size $T \times F$, where $T$ represents the number of features and $F$ is the dimensionality of each feature. To efficiently map the grid coordinates, a spatial hash function $h(x)$, based on the approach of \cite{teschner2003optimized}, is utilized:

\vspace{-25pt}
\begin{align} h(x) &= \left(\bigoplus_{i=1}^{d} x_{i} \pi_{i} \right) \text{ mod } T \end{align}
\vspace{-10pt}

Here, $\oplus$ denotes the bitwise XOR operation, and $\pi_{i}$ are distinct large prime numbers.

\subsection{Continual Neural Graphics Primitive (C-NGP)}

This section introduces the C-NGP framework that incorporates scene-conditioning and continual learning in the Instant-NGP network \cite{muller2022instant}. 

\underline{\textit{Conditioning:}} To condition Instant-NGP, we use pseudo integer labels $C \in$ \{1, 2, 3, ...\} to differentiate scenes. Figure \ref{fig:framework} illustrates the information flow. To get the Neural Hash Encoding (NHE) of coordinates, we combine the scene coordinates ($x$, $y$, $z$) with the pseudo label ($C$). Additionally, the pseudo labels undergo sinusoidal positional encoding (PE) \cite{tancik2020fourier}, denoted as $\psi \in \mathbb{R}^{2\alpha}$, where $\alpha=2$ represents the number of frequency components used. The neural hash encoding (NHE) and encoded pseudo labels $(\psi)$ are concatenated and passed through a single-layer MLP to obtain density ($\sigma$) and scene features. The predicted scene feature is then used to compute the color for each input coordinate ($x$, $y$, $z$). Specifically, the scene feature is concatenated with the encoded viewing direction ($SHE$) and encoded pseudo label ($\psi$), forming the input to a two-layer MLP that predicts the final color values. The viewing direction is encoded using spherical harmonics encoding (SHE) \cite{ramamoorthi2001efficient}, further refining the rendering process. We visualize how neural hashing helps segregate scenes within a shared representation space upon scene-conditioning in the supplementary material. 


\underline{\textit{Continual learning:}} While conditioning helps non-conflicting scene representations, continual learning accommodates new, unseen scenes incrementally. Therefore, the proposed C-NGP involves training Instant-NGP (with scene conditioning) in a continual learning paradigm as follows. Each scene is encoded into conditional Instant-NGP using a pseudo-label. Before encoding a new scene, we first render all previously observed scenes using the camera parameters of the new scene. These rendered images are then added to the training set of the new scene, and the model parameters are updated accordingly. This strategy eliminates reliance on previous training datasets while ensuring high-fidelity modeling of all observed scenes. Rendering previously seen scenes with updated camera parameters is particularly effective for datasets like NeRF Synthetic $360^{\circ}$. For other scene types, we use the stored camera parameters for rendering.

\subsection{Loss and Regularization}
Floaters represent a significant challenge in radiance fields, typically manifesting as disconnected, dense spatial regions near the camera plane. Therefore, apart from the conventional rendering loss $\mathcal{L}_{mse}$, we include two additional regularizations to ensure stable C-NGP training.

\textbf{Distortion.} Following the work of \cite{barron2022mip, wynn2023diffusionerf}, we used distortion loss for compact point distribution on the camera ray:

\vspace{-15pt}
\begin{align}
    \mathcal{L}_{dist} = \frac{1}{d(r)}\Bigg(\sum_{i}w_{i}w_{j} \Bigg|\frac{t_{i}+t_{i+1}}{2} - \frac{t_{j}+t_{j+1}}{2} \Bigg|
    + \frac{1}{3} \sum_{i=1}^{N} w_{i}^{2}(t_{i+1}-t_{i})\Bigg)
\end{align}

where, $d(r) = \frac{\sum_{i=1}^{N}w_{i}t_{i}}{\sum_{i=1}^{N}w_{i}}$ is depth along each ray. The first part of $\mathcal{L}_{dist}$ minimizes the weighted distance between all pairs of interval midpoints. The second part focuses on minimizing the weighted size of each interval. Jointly, the weights on the ray are encouraged to be compact by pulling distance intervals closer by consolidating each weight and minimizing the width of each interval \cite{barron2022mip}.

\textbf{Ray entropy.} The entropy regularization calculates the entropy of the ray's distribution \cite{kim2022infonerf, bonotto2024combinerf} for each ray passing through the scene. This involves assessing the uncertainty or randomness in the predicted densities along the ray and avoiding the floaters in the rendered scene. The entropy regularization is described as per Equation \ref{eq:ray_entropy} \vspace{-5pt}
\begin{align}
    \mathcal{L}_{ent} &= \left(-\sum_{i=1}^{N} p(r_{i}) log(p(r_{i})\right)
    \label{eq:ray_entropy}
\end{align}

Here, $N$ is the number of sampled points on the ray $r$, and $p_{i}$ is the opacity of each sampled point.

\textbf{Complete loss.} To train the C-NGP method, we used the combined loss formulation, as described in Equation \ref{eq:total_loss}.
\vspace{-8pt}
\begin{align}
    \mathcal{L}_{total} &= \mathcal{L}_{mse} + \lambda_{ent} \mathcal{L}_{ent} + \lambda_{dist} \mathcal{L}_{dist}
    \label{eq:total_loss}
\end{align}

Here, $\lambda_{ent}$ and $\lambda_{dist}$ are the weights for the regularization. We keep these parameters for training the complete network as $\lambda_{ent} = 1e-3$ and $\lambda_{dist}=1e-2$.
\section{Experiments}
\label{sec:experiment}
In this section, we perform an exhaustive set of experiments to demonstrate the efficacy of the proposed framework. We start by discussing the datasets, evaluation metrics, and comparison baselines.

\begin{figure}[!t]
  \centering
  \includegraphics[width=1.00\linewidth]{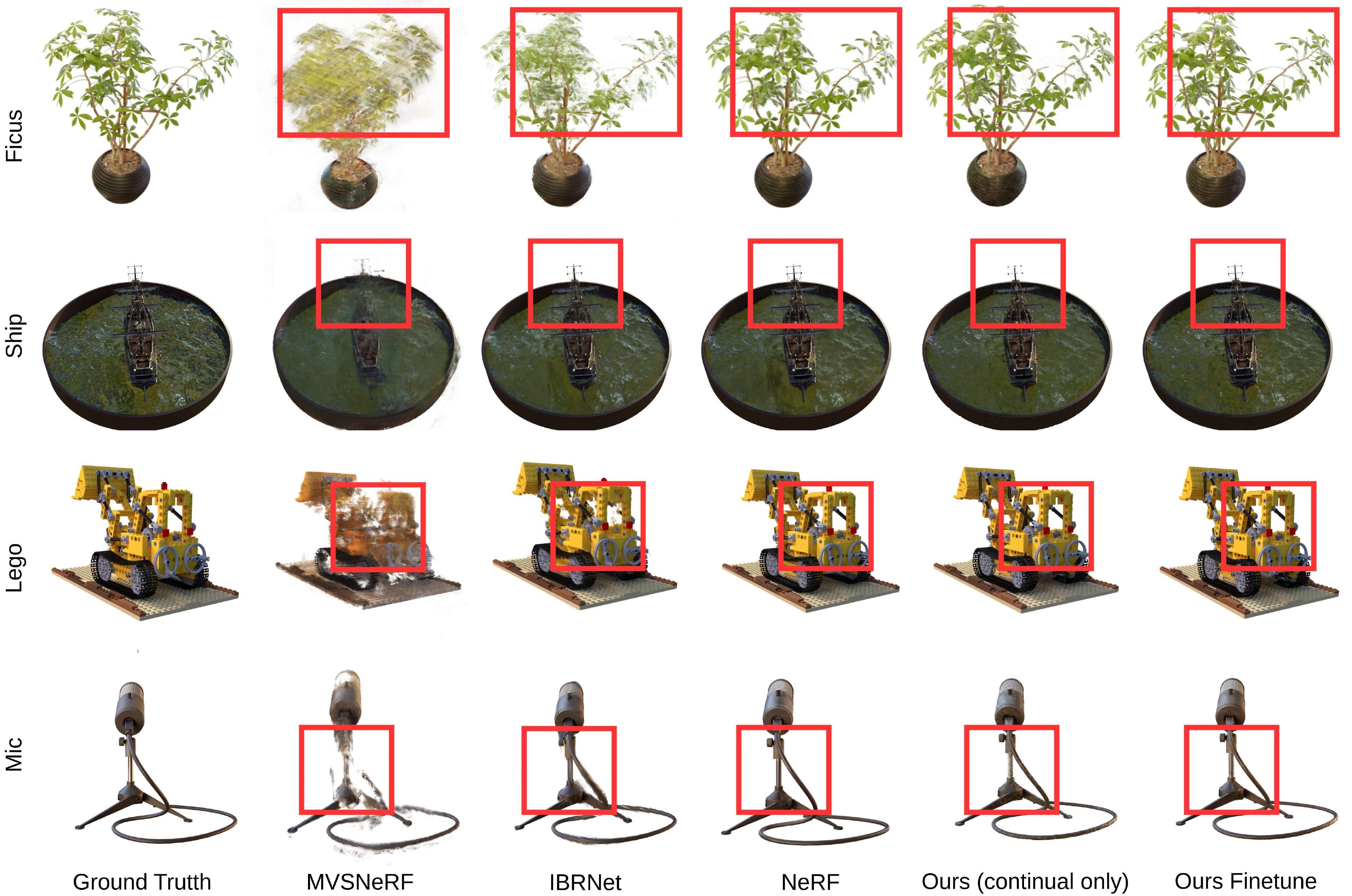}
  \vspace{0.1em}
  \caption{Qualitative Comparison on the Nerf Synthetic $360^{\circ}$ dataset \cite{mildenhall2021nerf}. C-NGP is compared with  MVSNeRF \cite{chen2021mvsnerf} and IBRNet \cite{wang2021ibrnet} (both with per-scene fine-tuning), and vanilla NeRF \cite{mildenhall2021nerf} with per-scene optimization. C-NGP is observed to model scenes with minimal artifacts compared to those present in the other methods. C-NGP (FT) is fine-tuned for each test scene for five epochs after continual learning.}
  \label{fig:qual_comparison_nerfsynth}
  \vspace{-0.5cm}
\end{figure}

\subsection{Datasets and Evaluation Metrics}
We evaluate our model on widely used datasets: NeRF Synthetic $360^{\circ}$ \cite{mildenhall2021nerf}, Tanks and Temples \cite{knapitsch2017tanks}, and Forward-Facing LLFF \cite{mildenhall2019local}. NeRF Synthetic $360^{\circ}$ consists of eight diverse scenes, each with $100$ training views and $200$ test views at a resolution of $800 \times 800$. The images are rendered from either an upper hemisphere or a full hemisphere, and the camera parameters are not necessarily consistent across scenes. Forward-Facing LLFF contains eight real-world scenes captured with a handheld camera in a forward-facing manner. The number of images per scene varies from $20$ to $62$, with a resolution of $1008 \times 756$. Tanks and Temples feature five large-scale scenes with complex geometries and real-world objects, each captured at a resolution of $1920 \times 1080$. We use the masked version of this dataset for training and testing, following \cite{liu2020neural}. Additionally, we generate a dataset of $22$ scenes using BlenderNeRF \cite{githubGitHubMaximeraafatBlenderNeRF}. Each scene is created from freely available 3D object meshes and rendered with $100$ training views and $100$ test views from an upper hemisphere. This dataset is designed to stress test the continual learning setup of C-NGP and analyze the representative upper bound of network parameters. To ensure fair evaluation in a continual learning setup, all scenes in this dataset are rendered with the same camera parameters.

\textbf{Metrics.} We quantitatively evaluate the model performance Peak-Signal-to-Noise-Ratio (PSNR) \cite{psnrmetric}, Structural Similarity Index (SSIM) \cite{wang2004image}, and Perceptual Score (LPIPS) \cite{zhang2018unreasonable}.

\subsection{Training Details}

For training C-NGP, we used a batch size of $10,000$ rays and an initial learning rate of $2\times 10^{-3}$. The model was trained for $30$ epochs across synthetic and real datasets. C-NGP was conditioned on pseudo labels assigned to each scene to ensure scene-specific representations. The hash table size for Instant-NGP was set to $T = 2^{19}$, with a feature size of $F=4$. For the NeRF Synthetic $360^{\circ}$ dataset, we applied the generative replay method \cite{shin2017continual}, where previously observed scenes were re-rendered using the camera parameters of the new scene. For real-world datasets like LLFF \cite{mildenhall2019local} and Tanks and Temples \cite{knapitsch2017tanks}, which lack full $360^{\circ}$ coverage, we only store the camera parameters of each scene. This allowed us to generate novel views while avoiding artifacts such as floaters or ghosting, less problematic issues in synthetic datasets with complete $360^{\circ}$ coverage.

\begin{figure}[!t]
  \centering
  \includegraphics[width=1.0\linewidth]{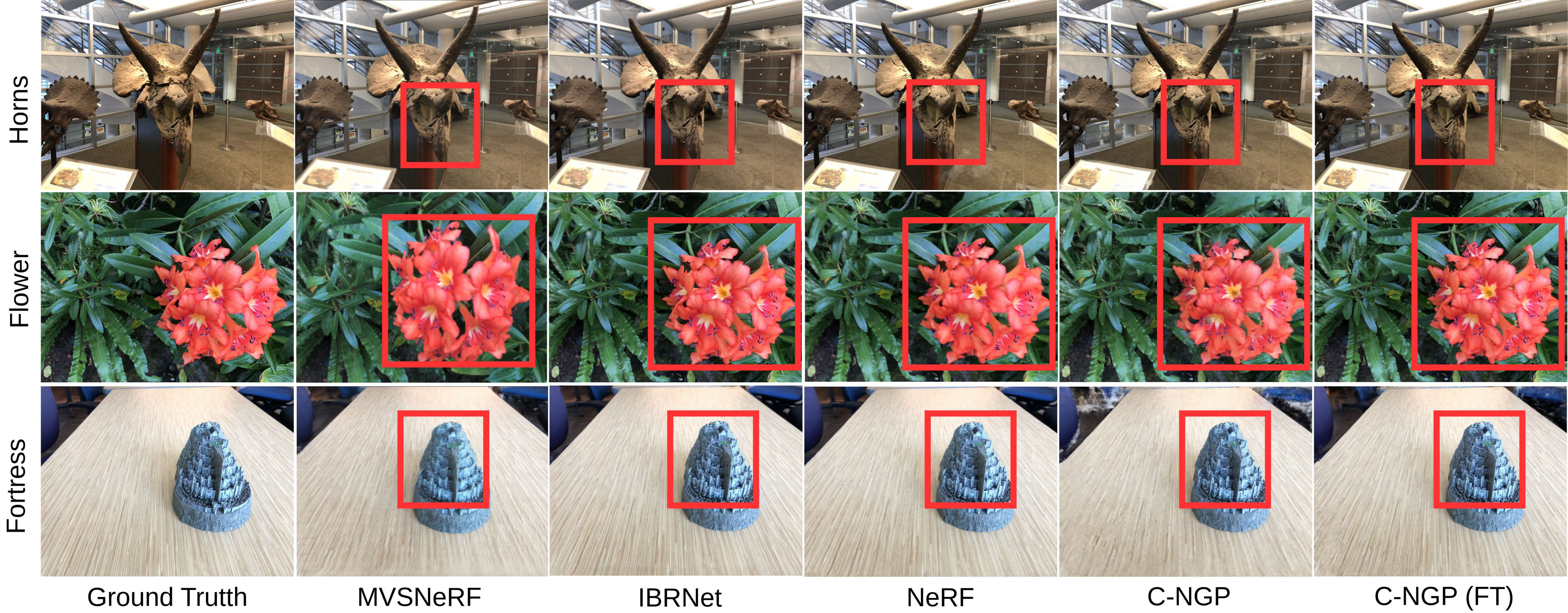} \vspace{0.1em}
  \caption{Qualitative Comparison on the Real Forward Facing (LLFF) dataset \cite{mildenhall2019local}. We compare C-NGP with  MVSNeRF \cite{chen2021mvsnerf} and IBRNet \cite{wang2021ibrnet} (both with per-scene fine-tuning), and vanilla NeRF \cite{mildenhall2021nerf} with per-scene. C-NGP (FT) is fine-tuned for each test scene for five epochs after continual learning.}
  \label{fig:qual_comparison_llff}
  \vspace{-2mm}
\end{figure}

\begin{table}[!t]
\centering
\begin{minipage}{0.48\textwidth}
\centering
{\small
\resizebox{0.85\linewidth}{!}{
\def\arraystretch{1.0}
\begin{tabular}{c|cc}
\hline
\multirow{2}{*}{Method}  & \multicolumn{2}{c}{Tanks and Temples} \\ \cline{2-3}
& PSNR $\uparrow$              & SSIM  $\uparrow$           \\ \hline
\multicolumn{1}{c|}{EWC \cite{kirkpatrick2017overcoming} + NeRF }  & 15.64             & 0.420            \\
\multicolumn{1}{c|}{PackNet \cite{mallya2018packnet} + NeRF}                                                                                      & 16.71             & 0.547            \\
\multicolumn{1}{c|}{MEIL-NeRF \cite{chung2022meil}}                                                                                           & 17.98             & 0.580            \\
\multicolumn{1}{c|}{CLNeRF \cite{shin2017continual}}                                                                                              & 21.30             & 0.640            \\
\multicolumn{1}{c|}{SCARF \cite{wang2024scarf}}                                                                                         & 26.78             & \textbf{0.89}    \\
\multicolumn{1}{c|}{{C-NGP (ours)}}                                                                                      & \textbf{26.93}    & {0.87}             \\ \hline
\end{tabular}}
}
\end{minipage}
\hfill
\begin{minipage}{0.48\textwidth}
\centering
{\small
\resizebox{\linewidth}{!}{
\def\arraystretch{1.5}

\begin{tabular}{c|ccc|ccc}
\hline
\multirow{2}{*}{Method} & \multicolumn{3}{c|}{NeRF Synthetic 360} & \multicolumn{3}{c}{Real Forward-Facing} \\ \cline{2-7} 
                        & PSNR $\uparrow$           & SSIM $\uparrow$        & LPIPS $\downarrow$ & PSNR $\uparrow$        & SSIM $\uparrow$      & LPIPS $\downarrow$      \\ \hline
SCARF \cite{wang2024scarf}                  & \textbf{30.94}  & 0.94          & -     & \textbf{26.44}        & \textbf{0.80}       & -           \\
C-NGP (ours)            & 29.40           & \textbf{0.94} & 0.09  & 22.12        & 0.66       & 0.37        \\ \hline
\end{tabular}
}}
{\small
\resizebox{\linewidth}{!}{
\def\arraystretch{1.5}
\begin{tabular}{c|ccc|ccc}
\hline
                                             & \multicolumn{3}{c|}{Offline Sampling}                                                         & \multicolumn{3}{c}{Online Sampling}                                                           \\ \cline{2-7} 
\multirow{-2}{*}{Datasets}                   & PSNR $\uparrow$                           & SSIM $\uparrow$                        & LPIPS $\downarrow$                         & PSNR  $\uparrow$                         & SSIM $\uparrow$                        & LPIPS $\downarrow$                        \\ \hline
\multicolumn{1}{c|}{NeRF Synthetic 360$^{\circ}$} & 25.211                         & 0.86                         & 0.151                         & 29.40                          & 0.93                         & 0.09                          \\
\multicolumn{1}{c|}{Blender Synthetic}   & 37.735 & 0.98 & 0.022 & 37.734 & 0.98 & 0.022 \\ \hline
\end{tabular}
}}
\end{minipage}
\vspace{1em}
\caption{(Left) Quantitative comparison of C-NGP on the Tanks and Temple dataset \cite{knapitsch2017tanks} with baselines that combine either conventional continual learning methods with NeRF or introduce continual learning for single scene optimization along with a concurrent work. (Right-Top) Quantitative comparison of C-NGP on NeRF Synthetic \cite{mildenhall2021nerf} and Real LLFF \cite{mildenhall2019local} with concurrent work. (Right-Bottom) Analyzing the quantitative performance on training with (online) and without (offline) generative replay while learning C-NGP. The enhanced performance with online sampling is due to the same camera parameters across all the scenes.}
\label{table:comparision_sampling}
\vspace{-3.0mm}
\end{table}


\subsection{Experimental Evaluation}

\textbf{Choice of Baselines.} Since no existing work incrementally models multiple scenes in a single radiance field, we compare our approach with the most closely related baselines that align, at least partially, with our experimental setup \cite{chen2021mvsnerf, wang2021ibrnet, he2023cp, wang2024scarf}. We have shown the quantitative analysis results in the supplementary. Among them, SCARF \cite{wang2024scarf} is the closest and most concurrent to our work. It handles multiple scenes, however, by increasing the number of parameters through a global parameter generator, unlike us. We compare the quantitative performance with SCARF based on their reported results in the paper across different datasets, and omit the qualitative comparison due to the unavailability of their code. 

\textbf{Quantitative.} We compare the proposed paradigm with continual learning frameworks like Elastic Weight Consolidation (EWC) and PacketNet \cite{mallya2018packnet} combined with NeRF, and single-scene continual NeRF methods like MEIL-NeRF \cite{chung2022meil} and CLNeRF \cite{shin2017continual}. As shown in Table \ref{table:comparision_sampling} (left), on the Tanks and Temples \cite{knapitsch2017tanks}, C-NGP achieves the highest PSNR and SSIM comparable to SCARF ($0.02$ units lower) while avoiding additional parameters per scene. On Nerf Synthetic $360^{\circ}$ \cite{mildenhall2021nerf} (see Table \ref{table:comparision_sampling} (right)), C-NGP shows competitive performance with SCARF. However, over scenes from the LLFF dataset \cite{mildenhall2019local}, C-NGP does not achieve the best scores due to the inherent struggle of Instant-NGP over the unbounded scenes. To support the claim, more empirical results are detailed in the supplementary material.

\textbf{Qualitative.} We provide a qualitative comparison of C-NGP with MVSNeRF \cite{chen2021mvsnerf}, IBRNet \cite{wang2021ibrnet}, and vanilla NeRF \cite{mildenhall2021nerf} in Figure \ref{fig:qual_comparison_nerfsynth} and \ref{fig:qual_comparison_llff}, over NeRF synthetic and Real LLFF datasets, respectively. Given its multi-scene modeling capability, we expect C-NGP’s performance to at least closely match per-scene optimized vanilla NeRF. Our results show that C-NGP achieves the best rendering quality on the NeRF Synthetic dataset and closely matches vanilla NeRF (which performs better than Instant-NGP) on the Real Forward-Facing dataset \cite{mildenhall2019local}. Additionally, we evaluate the visual fidelity of different scenes rendered by C-NGP across the NeRF Synthetic \cite{mildenhall2021nerf}, Real Forward-Facing \cite{mildenhall2019local}, and Tanks and Temples \cite{knapitsch2017tanks} in the supplementary. 

While C-NGP shows strong performance on synthetic datasets and Tanks \& Temples, its performance on unbounded LLFF scenes is relatively lower. We attribute this partly to the Instant-NGP backbone, which is known to be less effective in such settings \cite{barron2022mip,li2023nerfacc}. Nonetheless, C-NGP shows better performance than Instant-NGP after fine-tuning while retaining efficiency and scalability.

\begin{figure}[!t]
    \begin{minipage}[b]{\linewidth}
        \centering
        \resizebox{\linewidth}{!}{
            \begin{tabular}{cc}
             \includegraphics[width=0.65\linewidth]{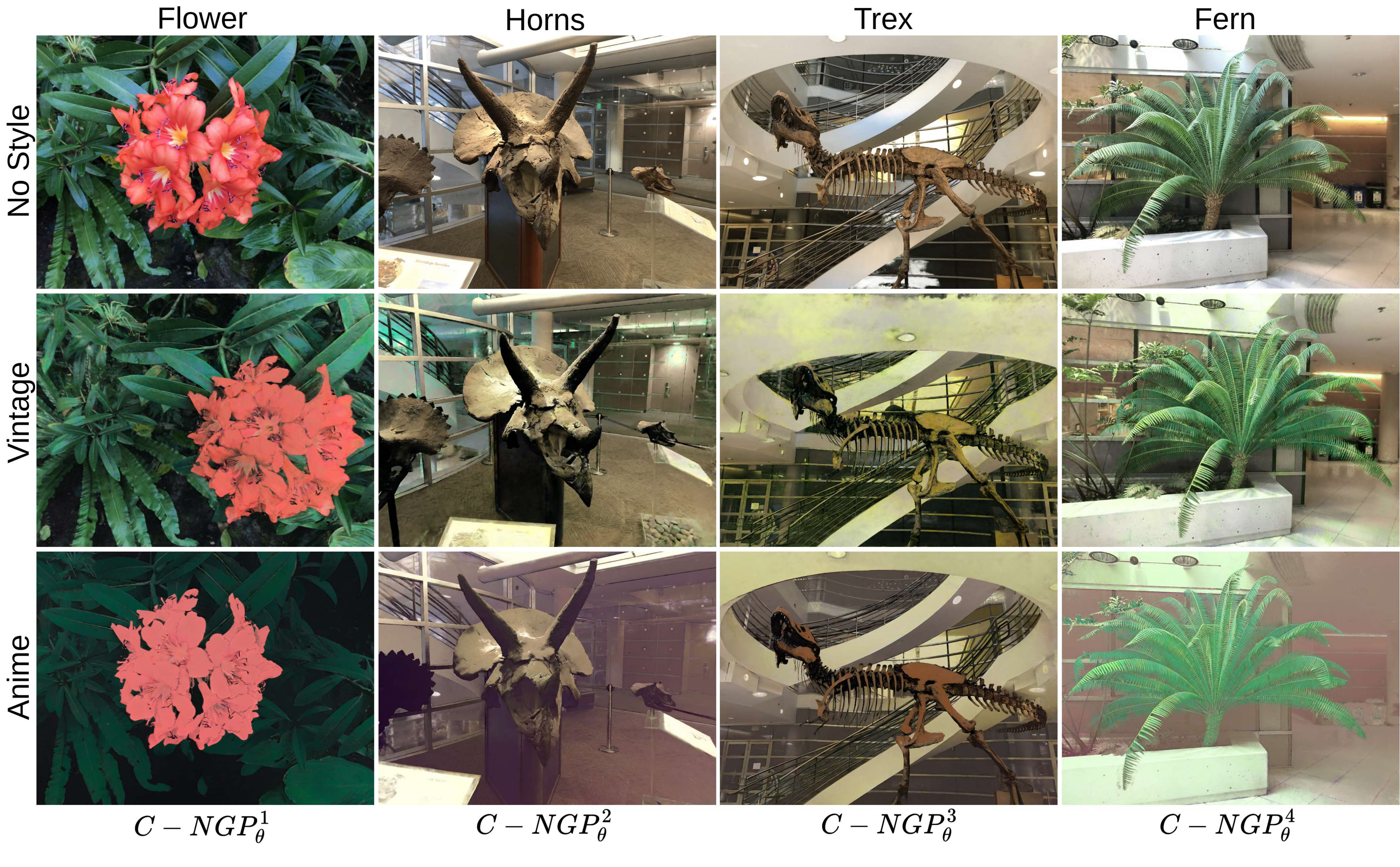}   &  \includegraphics[width=0.48\linewidth]{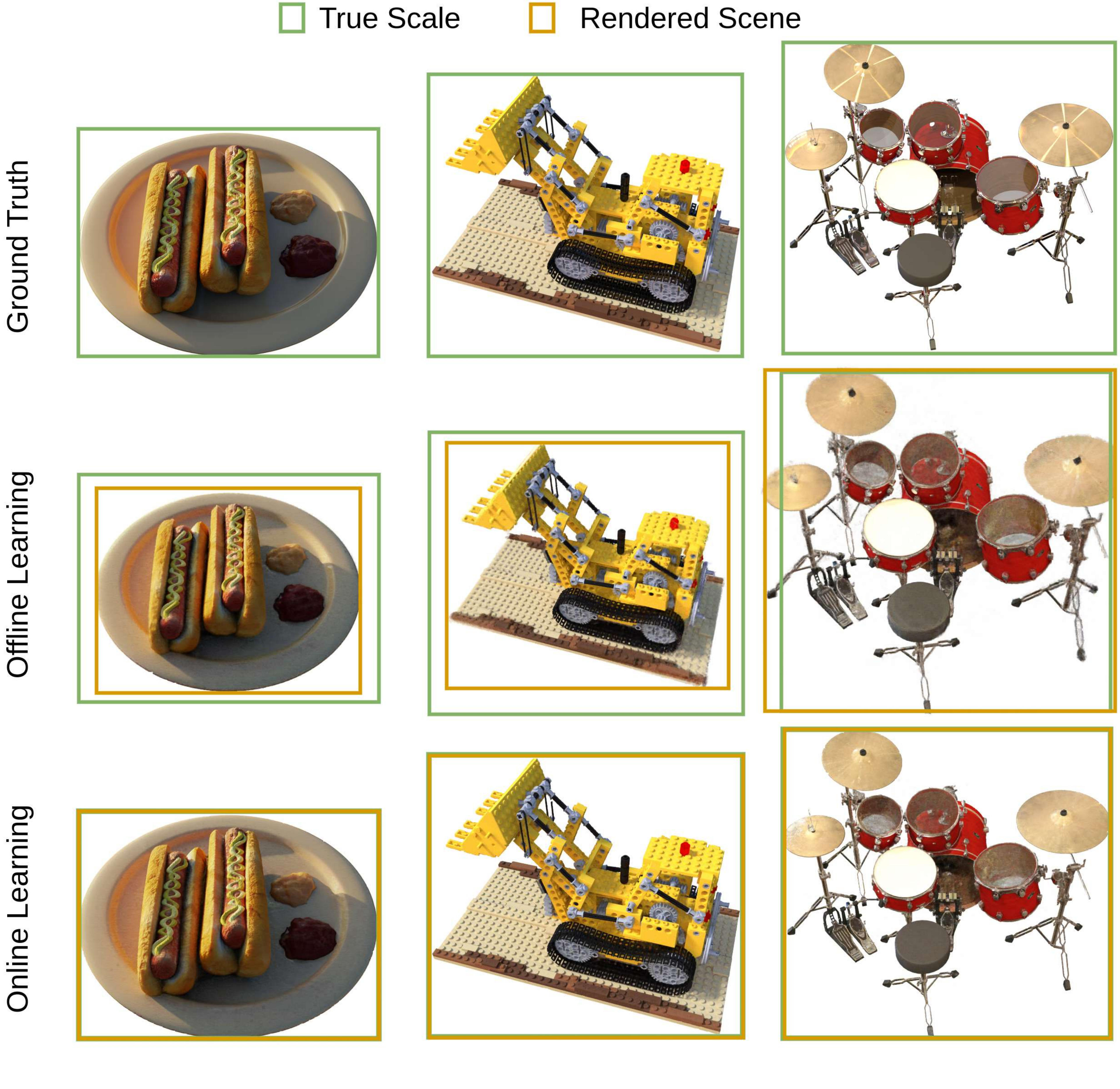}
            \end{tabular}
        }
        \vspace{0.5em}
        \captionof{figure}{(Left) The figure illustrates how the proposed C-NGP method stores multiple edit styles within a single network. Each scene is trained separately, represented by $\theta^{i}$, and stores different edit styles for respective scenes. (Right) Effect of the difference in camera parameters across scenes in the offline setting.}
        \label{fig:multistylengp}
    \end{minipage}
    \vspace{-0.75cm}
\end{figure}

\begin{table*}[!t]
\resizebox{\linewidth}{!}{
\def\arraystretch{1.5}%
\begin{tabular}{c|cccc|cccc|cccc|cccc}
\hline
\multirow{2}{*}{Method}                                                                            & \multicolumn{4}{c|}{Current Scene {[}\textbf{L}{]}} & \multicolumn{4}{c|}{Current Scene {[}L $\xrightarrow{}$ \textbf{C}{]}} & \multicolumn{4}{c|}{Current Scene {[}L $\xrightarrow{}$ C $\xrightarrow{}$ \textbf{S}{]}} & \multicolumn{4}{c}{Current Scene {[}L $\xrightarrow{}$ C $\xrightarrow{}$ S $\xrightarrow{}$ \textbf{D}{]}} \\ \cline{2-17} 
                                                                                                   & Lego       & Chair      & Ship     & Drums     & Lego       & Chair      & Ship     & Drums         & Lego       & Chair      & Ship     & Drums             & Lego       & Chair      & Ship     & Drums                   \\ \cline{1-17} 
\multicolumn{1}{c|}{\begin{tabular}[c]{@{}c@{}}Offline Sampling\\ \end{tabular}} & 35.62      & x      &   x        &    x      & 34.59           & 34.63           &       x          &        x       &          34.19             &           34.06             &          28.53               &         x              &             \cellcolor[HTML]{F4CCCC}14.41                 &   \cellcolor[HTML]{F4CCCC}18.12               &        28.06                   &    25.02                 \\ 
\multicolumn{1}{c|}{\begin{tabular}[c]{@{}c@{}}Online Sampling\\ \end{tabular}} & 35.62      & x     &   x        &    x      & 34.81           & 34.82           &       x          &        x       &          34.24             &           34.35             &         28.68               &         x              &             33.39                 &   33.89               &         28.29                   &    25.20                 \\ \hline
\end{tabular}}
\vspace{1em}
\caption{Analysing progressive degradation in the rendering quality with \textit{offline} and \textit{online sampling} of scenes from the  NeRF Synthetic $360^{\circ}$ dataset \cite{mildenhall2021nerf}. The cell highlighted in red color shows high degradation in the PSNR value under offline sampling.} 
\vspace{-0.1 cm}
\label{table:online_offline_deg}
\end{table*}

\subsection{Additional Results and Analysis}

\textbf{Style Editing.} To show the application of the proposed C-NGP framework, we adapt it for style editing in 3D scenes \cite{nguyen2022snerf, huang2021learning, kopanas2021point}. The current methods are only able to do single style transfer to any NeRF model at a time \cite{haque2023instruct, fujiwara2024style}, and for each new style, they have to retrain the NeRF model. The incremental nature of C-NGP also allows for accommodating multiple different styles within a scene. For each scene, we train a separate C-NGP network {$\theta^{1}$, $\theta^{2}$, $\theta^{3}$, $\theta^{4}$} and for each $\theta^{i}$, we first learn the same scene for $K$ times incrementally using C-NGP and to transfer the style, we only fine-tune the MLP layer responsible for color (RGB). As shown in Figure \ref{fig:multistylengp} (left), we learn two different styles for each scene, i.e., each $\theta^{i}$ stores three different variations of the same scene, where the first is the original scene itself, and the next two are the edited scene styles which are vintage and anime. To generate consistent multi-view style scenes for training, we follow the work by Fujiwara \textit{et al.} \cite{fujiwara2024style}, which uses fully fused attention and depth-conditioned ControlNet \cite{zhang2023adding}.


\begin{table}[!t]
\centering
\begin{minipage}{0.48\textwidth}
\centering
{\small
\resizebox{\linewidth}{!}{
\def\arraystretch{1.0}
\begin{tabular}{c|cc|cc}
\hline
                                          & \multicolumn{2}{c|}{Single Dataset} & \multicolumn{2}{c}{Mixed Dataset} \\ \cline{2-5} 
                                          & PSNR $\uparrow$             & SSIM $\uparrow$              & PSNR $\uparrow$               & SSIM $\uparrow$            \\ \hline
\multicolumn{1}{c|}{Groups 1 (8 Scenes)}  & 35.889             & 0.978             & 29.40               & 0.94              \\
\multicolumn{1}{c|}{Groups 2 (12 Scenes)} & 33.989             & 0.971             & 28.295              & 0.915             \\
\multicolumn{1}{c|}{Groups 3 (16 Scenes)} & 32.448             & 0.964             & 27.190              & 0.906             \\
\multicolumn{1}{c|}{Groups 4 (22 Scenes)} & 30.472             & 0.954             & 25.744              & 0.896             \\
 \hline
\end{tabular}
}}
\end{minipage}
\hfill
\begin{minipage}{0.48\textwidth}
\centering
{\small
\resizebox{\linewidth}{!}{
\def\arraystretch{1.0}
\begin{tabular}{c|ccc}
\hline
Experiments                              & PSNR $\uparrow$
 & SSIM $\uparrow$ & LPIPS $\downarrow$ \\ \hline
XYZ | C + $SE(\theta, \phi)$                                 & 24.108                        & 0.812                          & 0.201                            \\
XYZ + $\psi(C)$ + $SE(\theta, \phi)$                            & 12.966                        & 0.813                          & 0.460                            \\
XYZ | C + $\psi(C)$ + $SE(\theta, \phi)$          & 24.492                        & 0.824                          & 0.184                            \\ 
XYZ | C + $\psi(C)$ + $SE(\theta, \phi)$ + $\psi(C)$               & \cellcolor[HTML]{D0E0E3}\textbf{25.927} & \cellcolor[HTML]{D0E0E3}\textbf{0.880} & \cellcolor[HTML]{D0E0E3}\textbf{0.139} \\ \hline
\end{tabular}
}}
\end{minipage}
\vspace{1em}
\caption{(Left) Analysing upper bound on the number of scenes C-NGP can accommodate with reasonable loss in rendering quality across single and mixed datasets.  Single: $22$ scenes from BlenderNeRF dataset. Mixed: 14 scenes from BlenderNeRF + $8$ scenes from NeRF Synthetic $360^{\circ}$ dataset. (Right) Ablation on the configuration choice for conditioning Instant-NGP network \cite{muller2022instant} - the backbone of C-NGP.}
\label{table:group_and_ablation}
\vspace{-4mm}
\end{table}

\textbf{Sampling Strategy.} To study the effect of how well C-NGP performs when in a continual paradigm with and without using the generative replay method (using stored images of previous scenes). Here,  we define \textit{online sampling} and \textit{offline sampling} for the case with and without generative replay, respectively. As shown in Table \ref{table:comparision_sampling} (bottom-right), the performance with \textit{online} and \textit{offline} sampling is nearly the same over the BlenderNeRF dataset, in fact, better than with \textit{offline sampling} over the NeRF Synthetic dataset \cite{mildenhall2021nerf}. We attribute such performance trends to the nature of camera parameters. Figure \ref{fig:multistylengp} (right) shows such misalignments due to differences in camera parameters that eventually cause a reduction in PSNR, leading to reduced performance in the offline sampling. To further emphasize this, Table \ref{table:online_offline_deg} exclusively evaluates the rendering quality of scenes from the NeRF Synthetic dataset with these sampling methods.

\textbf{Upper bound.} Table \ref{table:group_and_ablation} (left) attempts to demonstrate the number of scenes that can be accommodated within a given set of parameters with an acceptable loss in the rendering quality of the previously learned scenes. While we do not claim to avoid forgetting completely, most importantly, we succeed in slowing it down to the extent that we can reasonably model around $\sim 20$ different scenes in a single neural radiance field.

\textbf{Design choices.} We evaluated different strategies for concatenating pseudo labels and their sinusoidal positional encodings with scene coordinates and viewing directions as summarized in Table \ref{table:group_and_ablation} (right). In our work, we use the configuration with the highest PSNR. More architectural design choices are detailed in the supplementary.

\textbf{Supplementary.} Due to space constraints, we elaborate: (a) quantitative comparison with other methods, (b) time analysis, (c) shared feature space visualization, and (d) additional discussion in the supplementary material. We urge the readers to kindly refer to the supplementary material as well for a holistic understanding of this work. These additional studies complement the main results and ensure that our evaluation is comprehensive, addressing performance, efficiency, and representation quality.

\section{Conclusion}
\label{sec:conclusion}

We introduced C-NGP, a continual learning framework for incremental multi-scene modeling within a single neural radiance field. The proposed method accommodates new scenes while reasonably preserving the previously learned information via scene conditioning and generative replay. Extensive empirical studies on synthetic and real datasets demonstrate that our method achieves strong performance in multi-scene representation learning and outperforms existing methods over training and inference speed, offering real-time rendering capabilities. C-NGP offers practical value for scalable 3D repositories, enabling multiple scenes and even style edits to be stored and shared in a single lightweight checkpoint, avoiding the overhead of per-scene models. While at this stage, the method incurs an acceptable loss in rendering quality while modeling multiple scenes, its performance in highly diverse scene settings and attempt towards zero forgetting warrants further exploration.

\section{Acknowledgment}
This work was supported by the Prime Minister Research Fellowship awarded to Prajwal Singh and Ashish Tiwari, and by the Jibaben Patel Chair in Artificial Intelligence held by Shanmuganathan Raman.

\bibliography{egbib}

\begin{thebibliography}{63}
\providecommand{\natexlab}[1]{#1}
\providecommand{\url}[1]{\texttt{#1}}
\expandafter\ifx\csname urlstyle\endcsname\relax
  \providecommand{\doi}[1]{doi: #1}\else
  \providecommand{\doi}{doi: \begingroup \urlstyle{rm}\Url}\fi

\bibitem[psn(1987)]{psnrmetric}
Implementation of a modified cvsd coder.
\newblock \emph{International Journal of Electronics}, 62\penalty0 (3):\penalty0 473--479, 1987.
\newblock \doi{10.1080/00207218708920998}.

\bibitem[Bao et~al.(2023)Bao, Zhang, Yang, Fan, Yang, Bao, Zhang, and Cui]{bao2023sine}
Chong Bao, Yinda Zhang, Bangbang Yang, Tianxing Fan, Zesong Yang, Hujun Bao, Guofeng Zhang, and Zhaopeng Cui.
\newblock Sine: Semantic-driven image-based nerf editing with prior-guided editing field.
\newblock In \emph{Proceedings of the IEEE/CVF Conference on Computer Vision and Pattern Recognition}, pages 20919--20929, 2023.

\bibitem[Barron et~al.(2022)Barron, Mildenhall, Verbin, Srinivasan, and Hedman]{barron2022mip}
Jonathan~T Barron, Ben Mildenhall, Dor Verbin, Pratul~P Srinivasan, and Peter Hedman.
\newblock Mip-nerf 360: Unbounded anti-aliased neural radiance fields.
\newblock In \emph{Proceedings of the IEEE/CVF Conference on Computer Vision and Pattern Recognition}, pages 5470--5479, 2022.

\bibitem[Bonotto et~al.(2024)Bonotto, Sarrocco, Evangelista, Imperoli, and Pretto]{bonotto2024combinerf}
Matteo Bonotto, Luigi Sarrocco, Daniele Evangelista, Marco Imperoli, and Alberto Pretto.
\newblock Combinerf: A combination of regularization techniques for few-shot neural radiance field view synthesis.
\newblock \emph{arXiv preprint arXiv:2403.14412}, 2024.

\bibitem[Cai and M{\"u}ller(2023)]{cai2023clnerf}
Zhipeng Cai and Matthias M{\"u}ller.
\newblock Clnerf: Continual learning meets nerf.
\newblock In \emph{Proceedings of the IEEE/CVF International Conference on Computer Vision}, pages 23185--23194, 2023.

\bibitem[Chen et~al.(2021)Chen, Xu, Zhao, Zhang, Xiang, Yu, and Su]{chen2021mvsnerf}
Anpei Chen, Zexiang Xu, Fuqiang Zhao, Xiaoshuai Zhang, Fanbo Xiang, Jingyi Yu, and Hao Su.
\newblock Mvsnerf: Fast generalizable radiance field reconstruction from multi-view stereo.
\newblock In \emph{Proceedings of the IEEE/CVF international conference on computer vision}, pages 14124--14133, 2021.

\bibitem[Chen et~al.(2022)Chen, Xu, Geiger, Yu, and Su]{chen2022tensorf}
Anpei Chen, Zexiang Xu, Andreas Geiger, Jingyi Yu, and Hao Su.
\newblock Tensorf: Tensorial radiance fields.
\newblock In \emph{European conference on computer vision}, pages 333--350. Springer, 2022.

\bibitem[Chen et~al.(2023{\natexlab{a}})Chen, Yi, Ma, Jia, and Lu]{chen2023gm}
Jianchuan Chen, Wentao Yi, Liqian Ma, Xu~Jia, and Huchuan Lu.
\newblock Gm-nerf: Learning generalizable model-based neural radiance fields from multi-view images.
\newblock In \emph{Proceedings of the IEEE/CVF Conference on Computer Vision and Pattern Recognition}, pages 20648--20658, 2023{\natexlab{a}}.

\bibitem[Chen et~al.(2023{\natexlab{b}})Chen, Chen, Wang, Zhang, Guo, Shan, and Wang]{chen2023local}
Yue Chen, Xingyu Chen, Xuan Wang, Qi~Zhang, Yu~Guo, Ying Shan, and Fei Wang.
\newblock Local-to-global registration for bundle-adjusting neural radiance fields.
\newblock In \emph{Proceedings of the IEEE/CVF Conference on Computer Vision and Pattern Recognition}, pages 8264--8273, 2023{\natexlab{b}}.

\bibitem[Chen et~al.(2024)Chen, Xu, Zheng, Zhuang, Pollefeys, Geiger, Cham, and Cai]{chen2024mvsplat}
Yuedong Chen, Haofei Xu, Chuanxia Zheng, Bohan Zhuang, Marc Pollefeys, Andreas Geiger, Tat-Jen Cham, and Jianfei Cai.
\newblock Mvsplat: Efficient 3d gaussian splatting from sparse multi-view images.
\newblock In \emph{European Conference on Computer Vision}, pages 370--386. Springer, 2024.

\bibitem[Chen et~al.(2023{\natexlab{c}})Chen, Funkhouser, Hedman, and Tagliasacchi]{chen2023mobilenerf}
Zhiqin Chen, Thomas Funkhouser, Peter Hedman, and Andrea Tagliasacchi.
\newblock Mobilenerf: Exploiting the polygon rasterization pipeline for efficient neural field rendering on mobile architectures.
\newblock In \emph{Proceedings of the IEEE/CVF Conference on Computer Vision and Pattern Recognition}, pages 16569--16578, 2023{\natexlab{c}}.

\bibitem[Chung et~al.(2022)Chung, Lee, Baik, and Lee]{chung2022meil}
Jaeyoung Chung, Kanggeon Lee, Sungyong Baik, and Kyoung~Mu Lee.
\newblock Meil-nerf: Memory-efficient incremental learning of neural radiance fields.
\newblock \emph{arXiv preprint arXiv:2212.08328}, 2022.

\bibitem[Deng et~al.(2024)Deng, Shen, Qin, Wang, Zhao, Wang, Wang, and Chen]{deng2024plgslam}
Tianchen Deng, Guole Shen, Tong Qin, Jianyu Wang, Wentao Zhao, Jingchuan Wang, Danwei Wang, and Weidong Chen.
\newblock Plgslam: Progressive neural scene representation with local to global bundle adjustment.
\newblock In \emph{Proceedings of the IEEE/CVF Conference on Computer Vision and Pattern Recognition}, pages 19657--19666, 2024.

\bibitem[Fang et~al.(2023)Fang, Xu, Wang, Yang, Wang, and Zhou]{fang2023one}
Shuangkang Fang, Weixin Xu, Heng Wang, Yi~Yang, Yufeng Wang, and Shuchang Zhou.
\newblock One is all: Bridging the gap between neural radiance fields architectures with progressive volume distillation.
\newblock In \emph{Proceedings of the AAAI Conference on Artificial Intelligence}, volume~37, pages 597--605, 2023.

\bibitem[Fridovich-Keil et~al.(2022)Fridovich-Keil, Yu, Tancik, Chen, Recht, and Kanazawa]{fridovich2022plenoxels}
Sara Fridovich-Keil, Alex Yu, Matthew Tancik, Qinhong Chen, Benjamin Recht, and Angjoo Kanazawa.
\newblock Plenoxels: Radiance fields without neural networks.
\newblock In \emph{Proceedings of the IEEE/CVF Conference on Computer Vision and Pattern Recognition}, pages 5501--5510, 2022.

\bibitem[Fujiwara et~al.(2024)Fujiwara, Mukuta, and Harada]{fujiwara2024style}
Haruo Fujiwara, Yusuke Mukuta, and Tatsuya Harada.
\newblock Style-nerf2nerf: 3d style transfer from style-aligned multi-view images.
\newblock In \emph{SIGGRAPH Asia 2024 Conference Papers}, pages 1--10, 2024.

\bibitem[Gordon et~al.(2023)Gordon, Chng, MacDonald, and Lucey]{gordon2023quantizing}
Cameron Gordon, Shin-Fang Chng, Lachlan MacDonald, and Simon Lucey.
\newblock On quantizing implicit neural representations.
\newblock In \emph{Proceedings of the IEEE/CVF Winter Conference on Applications of Computer Vision}, pages 341--350, 2023.

\bibitem[Haque et~al.(2023)Haque, Tancik, Efros, Holynski, and Kanazawa]{haque2023instruct}
Ayaan Haque, Matthew Tancik, Alexei~A Efros, Aleksander Holynski, and Angjoo Kanazawa.
\newblock Instruct-nerf2nerf: Editing 3d scenes with instructions.
\newblock In \emph{Proceedings of the IEEE/CVF International Conference on Computer Vision}, pages 19740--19750, 2023.

\bibitem[He et~al.(2023)He, Liang, Xiao, Chen, and Chen]{he2023cp}
Hao He, Yixun Liang, Shishi Xiao, Jierun Chen, and Yingcong Chen.
\newblock Cp-nerf: Conditionally parameterized neural radiance fields for cross-scene novel view synthesis.
\newblock In \emph{Computer Graphics Forum}, volume~42, page e14940. Wiley Online Library, 2023.

\bibitem[Huang et~al.(2021)Huang, Tseng, Saini, Singh, and Yang]{huang2021learning}
Hsin-Ping Huang, Hung-Yu Tseng, Saurabh Saini, Maneesh Singh, and Ming-Hsuan Yang.
\newblock Learning to stylize novel views.
\newblock In \emph{Proceedings of the IEEE/CVF International Conference on Computer Vision}, pages 13869--13878, 2021.

\bibitem[Kerbl et~al.(2023)Kerbl, Kopanas, Leimkuehler, and Drettakis]{Kerbl20233D}
Bernhard Kerbl, Georgios Kopanas, Thomas Leimkuehler, and George Drettakis.
\newblock {3D Gaussian Splatting for Real-Time Radiance Field Rendering}.
\newblock \emph{ACM Transactions on Graphics}, 2023.
\newblock \doi{10.1145/3592433}.

\bibitem[Kim et~al.(2022)Kim, Seo, and Han]{kim2022infonerf}
Mijeong Kim, Seonguk Seo, and Bohyung Han.
\newblock Infonerf: Ray entropy minimization for few-shot neural volume rendering.
\newblock In \emph{Proceedings of the IEEE/CVF Conference on Computer Vision and Pattern Recognition}, pages 12912--12921, 2022.

\bibitem[Kirkpatrick et~al.(2017)Kirkpatrick, Pascanu, Rabinowitz, Veness, Desjardins, Rusu, Milan, Quan, Ramalho, Grabska-Barwinska, et~al.]{kirkpatrick2017overcoming}
James Kirkpatrick, Razvan Pascanu, Neil Rabinowitz, Joel Veness, Guillaume Desjardins, Andrei~A Rusu, Kieran Milan, John Quan, Tiago Ramalho, Agnieszka Grabska-Barwinska, et~al.
\newblock Overcoming catastrophic forgetting in neural networks.
\newblock \emph{Proceedings of the national academy of sciences}, 114\penalty0 (13):\penalty0 3521--3526, 2017.

\bibitem[Knapitsch et~al.(2017)Knapitsch, Park, Zhou, and Koltun]{knapitsch2017tanks}
Arno Knapitsch, Jaesik Park, Qian-Yi Zhou, and Vladlen Koltun.
\newblock Tanks and temples: Benchmarking large-scale scene reconstruction.
\newblock \emph{ACM Transactions on Graphics (ToG)}, 36\penalty0 (4):\penalty0 1--13, 2017.

\bibitem[Kopanas et~al.(2021)Kopanas, Philip, Leimk{\"u}hler, and Drettakis]{kopanas2021point}
Georgios Kopanas, Julien Philip, Thomas Leimk{\"u}hler, and George Drettakis.
\newblock Point-based neural rendering with per-view optimization.
\newblock In \emph{Computer Graphics Forum}, volume~40, pages 29--43. Wiley Online Library, 2021.

\bibitem[Lee et~al.(2024)Lee, Choi, Kim, Kim, and Cho]{lee2024few}
SeokYeong Lee, JunYong Choi, Seungryong Kim, Ig-Jae Kim, and Junghyun Cho.
\newblock Few-shot neural radiance fields under unconstrained illumination.
\newblock In \emph{Proceedings of the AAAI Conference on Artificial Intelligence}, volume~38, pages 2938--2946, 2024.

\bibitem[Li et~al.(2023)Li, Gao, Tancik, and Kanazawa]{li2023nerfacc}
Ruilong Li, Hang Gao, Matthew Tancik, and Angjoo Kanazawa.
\newblock Nerfacc: Efficient sampling accelerates nerfs.
\newblock In \emph{Proceedings of the IEEE/CVF international conference on computer vision}, pages 18537--18546, 2023.

\bibitem[Li and Hoiem(2017)]{li2017learning}
Zhizhong Li and Derek Hoiem.
\newblock Learning without forgetting.
\newblock \emph{IEEE transactions on pattern analysis and machine intelligence}, 40\penalty0 (12):\penalty0 2935--2947, 2017.

\bibitem[Liu et~al.(2020)Liu, Gu, Zaw~Lin, Chua, and Theobalt]{liu2020neural}
Lingjie Liu, Jiatao Gu, Kyaw Zaw~Lin, Tat-Seng Chua, and Christian Theobalt.
\newblock Neural sparse voxel fields.
\newblock \emph{Advances in Neural Information Processing Systems}, 33:\penalty0 15651--15663, 2020.

\bibitem[Mallya and Lazebnik(2018)]{mallya2018packnet}
Arun Mallya and Svetlana Lazebnik.
\newblock Packnet: Adding multiple tasks to a single network by iterative pruning.
\newblock In \emph{Proceedings of the IEEE conference on Computer Vision and Pattern Recognition}, pages 7765--7773, 2018.

\bibitem[Max(1995)]{max1995optical}
Nelson Max.
\newblock Optical models for direct volume rendering.
\newblock \emph{IEEE Transactions on Visualization and Computer Graphics}, 1\penalty0 (2):\penalty0 99--108, 1995.

\bibitem[Mildenhall et~al.(2019)Mildenhall, Srinivasan, Ortiz-Cayon, Kalantari, Ramamoorthi, Ng, and Kar]{mildenhall2019local}
Ben Mildenhall, Pratul~P Srinivasan, Rodrigo Ortiz-Cayon, Nima~Khademi Kalantari, Ravi Ramamoorthi, Ren Ng, and Abhishek Kar.
\newblock Local light field fusion: Practical view synthesis with prescriptive sampling guidelines.
\newblock \emph{ACM Transactions on Graphics (ToG)}, 38\penalty0 (4):\penalty0 1--14, 2019.

\bibitem[Mildenhall et~al.(2021)Mildenhall, Srinivasan, Tancik, Barron, Ramamoorthi, and Ng]{mildenhall2021nerf}
Ben Mildenhall, Pratul~P Srinivasan, Matthew Tancik, Jonathan~T Barron, Ravi Ramamoorthi, and Ren Ng.
\newblock Nerf: Representing scenes as neural radiance fields for view synthesis.
\newblock \emph{Communications of the ACM}, 65\penalty0 (1):\penalty0 99--106, 2021.

\bibitem[M{\"u}ller et~al.(2022)M{\"u}ller, Evans, Schied, and Keller]{muller2022instant}
Thomas M{\"u}ller, Alex Evans, Christoph Schied, and Alexander Keller.
\newblock Instant neural graphics primitives with a multiresolution hash encoding.
\newblock \emph{ACM transactions on graphics (TOG)}, 41\penalty0 (4):\penalty0 1--15, 2022.

\bibitem[Nguyen-Phuoc et~al.(2022)Nguyen-Phuoc, Liu, and Xiao]{nguyen2022snerf}
Thu Nguyen-Phuoc, Feng Liu, and Lei Xiao.
\newblock Snerf: stylized neural implicit representations for 3d scenes.
\newblock \emph{arXiv preprint arXiv:2207.02363}, 2022.

\bibitem[Park et~al.(2019)Park, Florence, Straub, Newcombe, and Lovegrove]{park2019deepsdf}
Jeong~Joon Park, Peter Florence, Julian Straub, Richard Newcombe, and Steven Lovegrove.
\newblock Deepsdf: Learning continuous signed distance functions for shape representation.
\newblock In \emph{Proceedings of the IEEE/CVF conference on computer vision and pattern recognition}, pages 165--174, 2019.

\bibitem[Po et~al.(2023)Po, Dong, Bergman, and Wetzstein]{po2023instant}
Ryan Po, Zhengyang Dong, Alexander~W Bergman, and Gordon Wetzstein.
\newblock Instant continual learning of neural radiance fields.
\newblock In \emph{Proceedings of the IEEE/CVF International Conference on Computer Vision}, pages 3334--3344, 2023.

\bibitem[Raafat()]{githubGitHubMaximeraafatBlenderNeRF}
Maxime Raafat.
\newblock {G}it{H}ub - maximeraafat/{B}lender{N}e{R}{F}: {E}asy {N}e{R}{F} synthetic dataset creation within {B}lender --- github.com.
\newblock \url{https://github.com/maximeraafat/BlenderNeRF}.
\newblock [Accessed 19-05-2024].

\bibitem[Ramamoorthi and Hanrahan(2001)]{ramamoorthi2001efficient}
Ravi Ramamoorthi and Pat Hanrahan.
\newblock An efficient representation for irradiance environment maps.
\newblock In \emph{Proceedings of the 28th annual conference on Computer graphics and interactive techniques}, pages 497--500, 2001.

\bibitem[Rannen et~al.(2017)Rannen, Aljundi, Blaschko, and Tuytelaars]{rannen2017encoder}
Amal Rannen, Rahaf Aljundi, Matthew~B Blaschko, and Tinne Tuytelaars.
\newblock Encoder based lifelong learning.
\newblock In \emph{Proceedings of the IEEE international conference on computer vision}, pages 1320--1328, 2017.

\bibitem[Robins(1995)]{robins1995catastrophic}
Anthony Robins.
\newblock Catastrophic forgetting, rehearsal and pseudorehearsal.
\newblock \emph{Connection Science}, 7\penalty0 (2):\penalty0 123--146, 1995.

\bibitem[Shin et~al.(2017)Shin, Lee, Kim, and Kim]{shin2017continual}
Hanul Shin, Jung~Kwon Lee, Jaehong Kim, and Jiwon Kim.
\newblock Continual learning with deep generative replay.
\newblock \emph{Advances in neural information processing systems}, 30, 2017.

\bibitem[Shin and Park(2024)]{shin2024binary}
Seungjoo Shin and Jaesik Park.
\newblock Binary radiance fields.
\newblock \emph{Advances in neural information processing systems}, 36, 2024.

\bibitem[Sucar et~al.(2021)Sucar, Liu, Ortiz, and Davison]{sucar2021imap}
Edgar Sucar, Shikun Liu, Joseph Ortiz, and Andrew~J Davison.
\newblock imap: Implicit mapping and positioning in real-time.
\newblock In \emph{Proceedings of the IEEE/CVF international conference on computer vision}, pages 6229--6238, 2021.

\bibitem[Tancik et~al.(2020)Tancik, Srinivasan, Mildenhall, Fridovich-Keil, Raghavan, Singhal, Ramamoorthi, Barron, and Ng]{tancik2020fourier}
Matthew Tancik, Pratul Srinivasan, Ben Mildenhall, Sara Fridovich-Keil, Nithin Raghavan, Utkarsh Singhal, Ravi Ramamoorthi, Jonathan Barron, and Ren Ng.
\newblock Fourier features let networks learn high frequency functions in low dimensional domains.
\newblock \emph{Advances in neural information processing systems}, 33:\penalty0 7537--7547, 2020.

\bibitem[Tang et~al.(2022)Tang, Chen, Wang, and Zeng]{tang2022compressible}
Jiaxiang Tang, Xiaokang Chen, Jingbo Wang, and Gang Zeng.
\newblock Compressible-composable nerf via rank-residual decomposition.
\newblock \emph{Advances in Neural Information Processing Systems}, 35:\penalty0 14798--14809, 2022.

\bibitem[Teschner et~al.(2003)Teschner, Heidelberger, M{\"u}ller, Pomerantes, and Gross]{teschner2003optimized}
Matthias Teschner, Bruno Heidelberger, Matthias M{\"u}ller, Danat Pomerantes, and Markus~H Gross.
\newblock Optimized spatial hashing for collision detection of deformable objects.
\newblock In \emph{Vmv}, volume~3, pages 47--54, 2003.

\bibitem[Van~der Maaten and Hinton(2008)]{van2008visualizing}
Laurens Van~der Maaten and Geoffrey Hinton.
\newblock Visualizing data using t-sne.
\newblock \emph{JMLR}, 9\penalty0 (11), 2008.

\bibitem[Wan et~al.(2023)Wan, Richardt, Bo{\v{z}}i{\v{c}}, Li, Rengarajan, Nam, Xiang, Li, Zhu, Ranjan, et~al.]{wan2023learning}
Ziyu Wan, Christian Richardt, Alja{\v{z}} Bo{\v{z}}i{\v{c}}, Chao Li, Vijay Rengarajan, Seonghyeon Nam, Xiaoyu Xiang, Tuotuo Li, Bo~Zhu, Rakesh Ranjan, et~al.
\newblock Learning neural duplex radiance fields for real-time view synthesis.
\newblock In \emph{Proceedings of the IEEE/CVF Conference on Computer Vision and Pattern Recognition}, pages 8307--8316, 2023.

\bibitem[Wang et~al.(2023{\natexlab{a}})Wang, Chen, Loy, and Liu]{wang2023sparsenerf}
Guangcong Wang, Zhaoxi Chen, Chen~Change Loy, and Ziwei Liu.
\newblock Sparsenerf: Distilling depth ranking for few-shot novel view synthesis.
\newblock In \emph{Proceedings of the IEEE/CVF International Conference on Computer Vision}, pages 9065--9076, 2023{\natexlab{a}}.

\bibitem[Wang et~al.(2021{\natexlab{a}})Wang, Liu, Liu, Theobalt, Komura, and Wang]{wang2021neus}
Peng Wang, Lingjie Liu, Yuan Liu, Christian Theobalt, Taku Komura, and Wenping Wang.
\newblock Neus: Learning neural implicit surfaces by volume rendering for multi-view reconstruction.
\newblock \emph{arXiv preprint arXiv:2106.10689}, 2021{\natexlab{a}}.

\bibitem[Wang et~al.(2021{\natexlab{b}})Wang, Wang, Genova, Srinivasan, Zhou, Barron, Martin-Brualla, Snavely, and Funkhouser]{wang2021ibrnet}
Qianqian Wang, Zhicheng Wang, Kyle Genova, Pratul~P Srinivasan, Howard Zhou, Jonathan~T Barron, Ricardo Martin-Brualla, Noah Snavely, and Thomas Funkhouser.
\newblock Ibrnet: Learning multi-view image-based rendering.
\newblock In \emph{Proceedings of the IEEE/CVF Conference on Computer Vision and Pattern Recognition}, pages 4690--4699, 2021{\natexlab{b}}.

\bibitem[Wang et~al.(2023{\natexlab{b}})Wang, Wang, Qu, and Qi]{wang2023rip}
Yuze Wang, Junyi Wang, Yansong Qu, and Yue Qi.
\newblock Rip-nerf: learning rotation-invariant point-based neural radiance field for fine-grained editing and compositing.
\newblock In \emph{Proceedings of the 2023 ACM International Conference on Multimedia Retrieval}, pages 125--134, 2023{\natexlab{b}}.

\bibitem[Wang et~al.(2024)Wang, Wang, Wang, Duan, Bao, and Qi]{wang2024scarf}
Yuze Wang, Junyi Wang, Chen Wang, Wantong Duan, Yongtang Bao, and Yue Qi.
\newblock Scarf: Scalable continual learning framework for memory-efficient multiple neural radiance fields.
\newblock \emph{arXiv preprint arXiv:2409.04482}, 2024.

\bibitem[Wang et~al.(2004)Wang, Bovik, Sheikh, and Simoncelli]{wang2004image}
Zhou Wang, Alan~C Bovik, Hamid~R Sheikh, and Eero~P Simoncelli.
\newblock Image quality assessment: from error visibility to structural similarity.
\newblock \emph{IEEE transactions on image processing}, 13\penalty0 (4):\penalty0 600--612, 2004.

\bibitem[Wynn and Turmukhambetov(2023)]{wynn2023diffusionerf}
Jamie Wynn and Daniyar Turmukhambetov.
\newblock Diffusionerf: Regularizing neural radiance fields with denoising diffusion models.
\newblock In \emph{Proceedings of the IEEE/CVF Conference on Computer Vision and Pattern Recognition}, pages 4180--4189, 2023.

\bibitem[Yang et~al.(2023{\natexlab{a}})Yang, Hong, Li, Hu, Li, Lee, and Wang]{yang2023contranerf}
Hao Yang, Lanqing Hong, Aoxue Li, Tianyang Hu, Zhenguo Li, Gim~Hee Lee, and Liwei Wang.
\newblock Contranerf: Generalizable neural radiance fields for synthetic-to-real novel view synthesis via contrastive learning.
\newblock In \emph{Proceedings of the IEEE/CVF Conference on Computer Vision and Pattern Recognition}, pages 16508--16517, 2023{\natexlab{a}}.

\bibitem[Yang et~al.(2023{\natexlab{b}})Yang, Pavone, and Wang]{yang2023freenerf}
Jiawei Yang, Marco Pavone, and Yue Wang.
\newblock Freenerf: Improving few-shot neural rendering with free frequency regularization.
\newblock In \emph{Proceedings of the IEEE/CVF conference on computer vision and pattern recognition}, pages 8254--8263, 2023{\natexlab{b}}.

\bibitem[Yu et~al.(2021)Yu, Ye, Tancik, and Kanazawa]{yu2021pixelnerf}
Alex Yu, Vickie Ye, Matthew Tancik, and Angjoo Kanazawa.
\newblock pixelnerf: Neural radiance fields from one or few images.
\newblock In \emph{Proceedings of the IEEE/CVF Conference on Computer Vision and Pattern Recognition}, pages 4578--4587, 2021.

\bibitem[Zhang et~al.(2023{\natexlab{a}})Zhang, Li, Chen, and Xu]{zhang2023nerf}
Letian Zhang, Ming Li, Chen Chen, and Jie Xu.
\newblock Il-nerf: Incremental learning for neural radiance fields with camera pose alignment.
\newblock \emph{arXiv preprint arXiv:2312.05748}, 2023{\natexlab{a}}.

\bibitem[Zhang et~al.(2023{\natexlab{b}})Zhang, Rao, and Agrawala]{zhang2023adding}
Lvmin Zhang, Anyi Rao, and Maneesh Agrawala.
\newblock Adding conditional control to text-to-image diffusion models.
\newblock In \emph{Proceedings of the IEEE/CVF international conference on computer vision}, pages 3836--3847, 2023{\natexlab{b}}.

\bibitem[Zhang et~al.(2018)Zhang, Isola, Efros, Shechtman, and Wang]{zhang2018unreasonable}
Richard Zhang, Phillip Isola, Alexei~A Efros, Eli Shechtman, and Oliver Wang.
\newblock The unreasonable effectiveness of deep features as a perceptual metric.
\newblock In \emph{Proceedings of the IEEE conference on computer vision and pattern recognition}, pages 586--595, 2018.

\bibitem[Zhang et~al.(2022)Zhang, Bi, Sunkavalli, Su, and Xu]{zhang2022nerfusion}
Xiaoshuai Zhang, Sai Bi, Kalyan Sunkavalli, Hao Su, and Zexiang Xu.
\newblock Nerfusion: Fusing radiance fields for large-scale scene reconstruction.
\newblock In \emph{Proceedings of the IEEE/CVF Conference on Computer Vision and Pattern Recognition}, pages 5449--5458, 2022.

\end{thebibliography}

\clearpage \appendix \setcounter{page}{1}

\maketitlesupplementary

\label{sec:appendix_section}

\section{Quantitative Comparison with Other Methods}

The baseline methods report performance under two experimental settings: \textit{with} and \textit{without finetuning}. \textit{(a) \underline{Without finetuning}}: The model is trained on a set of scenes and evaluated on test scenes without modifying the learned parameters. \textit{(b) \underline{With finetuning}}: The model is first trained on a set of scenes, then fine-tuned on training images of a specific scene for a few epochs before evaluation on its unseen views. Although finetuning may seem redundant, as the model has already seen those scenes and will naturally perform better, we introduce it for a fair comparison similar to CP-NeRF \cite{he2023cp}, which also reports performance after finetuning on specific scenes.

To ensure a fair comparison with C-NGP, we determine the best setting for each baseline. Since MVSNeRF \cite{chen2021mvsnerf} and IBRNet \cite{wang2021ibrnet} are not trained on scenes from NeRF Synthetic \cite{mildenhall2021nerf} or Real forward-facing \cite{mildenhall2019local}, we compare MVSNeRF and IBRNet \textit{with finetuning} over these datasets. In contrast, NeRF*, CP-NeRF \cite{he2023cp}, and C-NGP are optimized on scenes from these datasets and are evaluated \textit{without fine-tuning}.

We quantitatively compare C-NGP with the baselines under the appropriate settings on the NeRF Synthetic $360^{\circ}$ \cite{mildenhall2021nerf} and Real forward-facing (LLFF) \cite{mildenhall2019local} datasets, as shown in Table \ref{table:nerf_llff} (left) and Table \ref{table:nerf_llff} (right), respectively. On the NeRF Synthetic $360^{\circ}$ dataset, C-NGP achieves the best SSIM and LPIPS scores under both settings. In terms of PSNR, it falls short by only $0.14$ units in the no-finetuning setting despite learning continually without any additional network for conditioning. This minor trade-off comes with the significant advantage of modeling multiple scenes incrementally without expanding the model size.

On the Real forward-facing dataset, as shown in Table \ref{table:nerf_llff} (right), C-NGP does not achieve the highest performance. This is primarily due to the limitations of the Instant-NGP backbone, which itself struggles with forward-facing scenes due to unbounded scenes, as observed when compared to vanilla NeRF \cite{mildenhall2021nerf} (see Table \ref{table:nerf_llff} (right), Row 4 \& 5). However, upon finetuning, C-NGP surpasses Instant-NGP (see Table \ref{table:nerf_llff} (right), Row 8). It is important to note that Instant-NGP was chosen as the backbone to leverage its faster training benefits, and we also evaluated vanilla NeRF in the proposed continual setting to demonstrate its capability to model multiple scenes.

\begin{table}[!t]
\centering
\begin{minipage}{0.48\textwidth}
\centering
{\small
\resizebox{\linewidth}{!}{
\def\arraystretch{1.0}
\begin{tabular}{c|c|ccc}
\toprule
& & \multicolumn{3}{c}{NeRF Synthetic 360$^{\circ}$} \\
\cmidrule(lr){3-5}
\multirow{-2}{*}{Method} & \multirow{-2}{*}{Finetuning} & PSNR $\uparrow$ & SSIM $\uparrow$ & LPIPS $\downarrow$ \\
\midrule
NeRF* & x & 21.75 & 0.84 & 0.16 \\
IBRNet \cite{wang2021ibrnet} & \checkmark & 28.14 & 0.94 & 0.07 \\
MVSNeRF \cite{chen2021mvsnerf} & \checkmark & 27.07 & 0.93 & 0.17 \\
CP-NeRF \cite{he2023cp} & x & \textbf{29.54} & 0.92 & 0.09 \\
\textbf{C-NGP (ours)} & x & 29.40 & \textbf{0.94} & \textbf{0.09} \\
\midrule
CP-NeRF \cite{he2023cp} & \checkmark & 31.77 & 0.95 & 0.06 \\
\textbf{C-NGP (ours)} & \checkmark & \textbf{32.08} & \textbf{0.95} & \textbf{0.06} \\
\bottomrule
\end{tabular}}
}
\label{table:nerf_synth}
\end{minipage}
\hfill
\begin{minipage}{0.48\textwidth}
\centering
{\small
\resizebox{\linewidth}{!}{
\def\arraystretch{1.0}
\begin{tabular}{c|c|ccc}
\toprule
& & \multicolumn{3}{c}{Real Forward-Facing} \\
\cmidrule(lr){3-5}
\multirow{-2}{*}{Method} & \multirow{-2}{*}{Finetuning} & PSNR $\uparrow$ & SSIM $\uparrow$ & LPIPS $\downarrow$ \\
\midrule
IBRNet \cite{wang2021ibrnet} & \checkmark & 26.73 & 0.85 & 0.18 \\
MVSNeRF \cite{chen2021mvsnerf} & \checkmark & 25.45 & \textbf{0.88} & 0.19 \\
CP-NeRF \cite{he2023cp} & x & 25.41 & 0.77 & 0.20 \\
\midrule
NeRF \cite{mildenhall2021nerf} & per-scene opt. & 26.50 & 0.81 & 0.25 \\
Instant-NGP \cite{muller2022instant} & per-scene opt. & \cellcolor[HTML]{F4CCCC}24.98 & \cellcolor[HTML]{F4CCCC}0.78 & \cellcolor[HTML]{F4CCCC}0.24 \\
\textbf{C-NGP (ours)} & x & \cellcolor[HTML]{D0E0E3}22.12 & \cellcolor[HTML]{D0E0E3}0.66 & \cellcolor[HTML]{D0E0E3}0.37 \\
\midrule
CP-NeRF \cite{he2023cp} & \checkmark & \textbf{27.23} & 0.81 & \textbf{0.14} \\
\textbf{C-NGP (ours)} & \checkmark & \cellcolor[HTML]{D0E0E3}25.19 & \cellcolor[HTML]{D0E0E3}0.76 & \cellcolor[HTML]{D0E0E3}0.25 \\
\bottomrule
\end{tabular}
}}
\end{minipage}
\vspace{1em}
\caption{Quantitative comparison on NeRF Synthetic $360^{\circ}$ dataset \cite{mildenhall2021nerf} (left) and Real Forward-Facing dataset \cite{mildenhall2019local} (right). Left: Our continual framework shows competitive performance with multi-scene modeling and outperforms previous state-of-the-art methods. NeRF* indicates vanilla NeRF trained under the proposed continual setup. Right: We compare C-NGP with Instant-NGP (highlighted in red) and vanilla NeRF. C-NGP's performance is constrained by Instant-NGP but improves with fine-tuning.}
\label{table:nerf_llff}
\end{table}

\begin{table*}[!t]
\resizebox{\linewidth}{!}{
\def\arraystretch{1.5}%
\begin{tabular}{c|cccc|cccc|cccc|ccccc}
\hline
\multirow{2}{*}{Method}                                                                            & \multicolumn{4}{c|}{Per Scene {[}S, F, H, L{]}} & \multicolumn{4}{c|}{Fine-tune {[}S $\xrightarrow{}$ F{]}} & \multicolumn{4}{c|}{Fine-tune {[}S $\xrightarrow{}$ F $\xrightarrow{}$ H{]}} & \multicolumn{5}{c}{Fine-tune {[}S $\xrightarrow{}$ F $\xrightarrow{}$ H $\xrightarrow{}$ L{]}} \\ \cline{2-18} 
                                                                                                   & Ship       & Ficus      & Hotdog     & Lego     & Ship            & Ficus           & Hotdog          & Lego          & Ship                  & Ficus                  & Hotdog                  & Lego                  & Ship                         & Ficus                         & Hotdog                         & Lego                         & Avg.                         \\ \cline{1-18} 
\multicolumn{1}{c|}{MVSNeRF \cite{chen2021mvsnerf}}                               & 21.27      & 19.60      & 22.44       & 18.90    & 12.61           & 17.72           & x               & x             & 15.52                 & 16.83                 & 18.50                  & x                     & 14.63                        & 16.75                        & 15.80                         & 14.43                        & 15.40                            \\
\multicolumn{1}{c|}{IBRNet \cite{wang2021ibrnet}}                                 & 28.70      & 29.19      & 37.14       & 28.23    & 22.63           & 25.23           & x               & x             & 23.98                 & 24.63                 & 35.16                  & x                     & 23.97                        & 24.71                        & 33.49                         & 30.72                        & 28.22                            \\ \hline
\multicolumn{1}{c|}{\begin{tabular}[c]{@{}c@{}}C-NGP (\textbf{ours})\\(no fine-tuning)\\ \end{tabular}} & 30.14      & 33.95      & 37.38       & 35.62    & 28.86           & 32.03           & x               & x             & 28.56                 & 31.03                 & 35.31                  & x                     & 28.06                        & 30.61                        & 34.62                         & 32.31                        & \textbf{31.4}                            \\ \hline
\end{tabular}} \vspace{1.0em}
\caption{Comparing the extent of information forgetting in MVSNeRF \cite{chen2021mvsnerf}, IBRNet \cite{wang2021ibrnet}, and C-NGP across scenes in the NeRF Synthetic $360^{\circ}$ dataset \cite{mildenhall2021nerf}.}
\label{table:forgetting}
\end{table*}

\begin{table}[!t]
\centering
\resizebox{0.6\linewidth}{!}{
\def\arraystretch{1.1}
\begin{tabular}{c|c|c|c}
\hline
Method & \begin{tabular}[c]{@{}c@{}}Training Time/\\ Complete Dataset\end{tabular} & \begin{tabular}[c]{@{}c@{}}Fine-tuning Time/\\ Scene\end{tabular} & \begin{tabular}[c]{@{}c@{}}Rendering Time/\\ Frame\end{tabular} \\ \hline
NeRF* & $\sim$12 days & x & $\sim$30 sec \\
MVSNeRF \cite{chen2021mvsnerf} & x & $\sim$1 hour & $\sim$14 sec \\
IBRNet \cite{wang2021ibrnet} & x & $\sim$9 hours & $\sim$18 sec \\
CP-NeRF \cite{he2023cp} & \textgreater 2 days & x & x \\
C-NGP \textbf{(ours)} & \textbf{$\sim$8 hours} & \textbf{$\sim$10 mins} & \textbf{$\sim$1.2 sec} \\ \hline
\end{tabular}}\vspace{1.0em}
\caption{Training, fine-tuning, and rendering time comparison of C-NGP with other methods on the dataset NeRF Synthetic $360^{\circ}$. All time comparison experiments are done on \textsf{NVIDIA RTX Quadro 5000} GPU. NeRF* indicates vanilla NeRF trained under the proposed continual learning setup.}
\label{table:timecomp}
\vspace{-3mm}
\end{table}

Notably, if the training data of previously observed scenes is available, fine-tuning C-NGP on those scenes can further improve rendering quality. However, we see, even without finetuning, C-NGP delivers significantly better results than the baselines. 

\textbf{Forgetting Effect.} Interestingly, Table \ref{table:forgetting} compares the extent of forgetting in MVSNeRF \cite{chen2021mvsnerf} and IBRNet \cite{wang2021ibrnet} with C-NeRF (not fine-tuned) evaluated over the scenes in the NeRF Synthetic $360^{\circ}$ \cite{mildenhall2021nerf} dataset. The experimental setup for the forgetting experiment is that we first fine-tune a scene on MVSNeRF and IBRNet. Later, for each new scene, we use the previous scene's fine-tuned weights and fine-tune them. For the proposed method, we follow the continual learning strategy only. No separate fine-tuning is done for C-NGP.

\section{Time Analysis}
\label{sec:timeanalysis}

Training time and rendering speed are crucial factors when modeling multiple scenes. As shown in Table \ref{table:timecomp}, NeRF requires $12$ days to train and $30$ seconds per frame to render, making it impractical. Other methods like MVSNeRF \cite{chen2021mvsnerf} and IBRNet \cite{wang2021ibrnet} reduce training time but still have long fine-tuning and rendering times. In contrast, C-NGP trains in just $8$ hours, fine-tuning takes $10$ minutes per scene, and renders at $1.2$ seconds per frame, making it highly efficient for real-time applications. 

\begin{figure}[!t]
    \begin{minipage}[b]{\linewidth}
        \centering
        \resizebox{\linewidth}{!}{
            \begin{tabular}{cc}
             \includegraphics[width=0.65\linewidth]{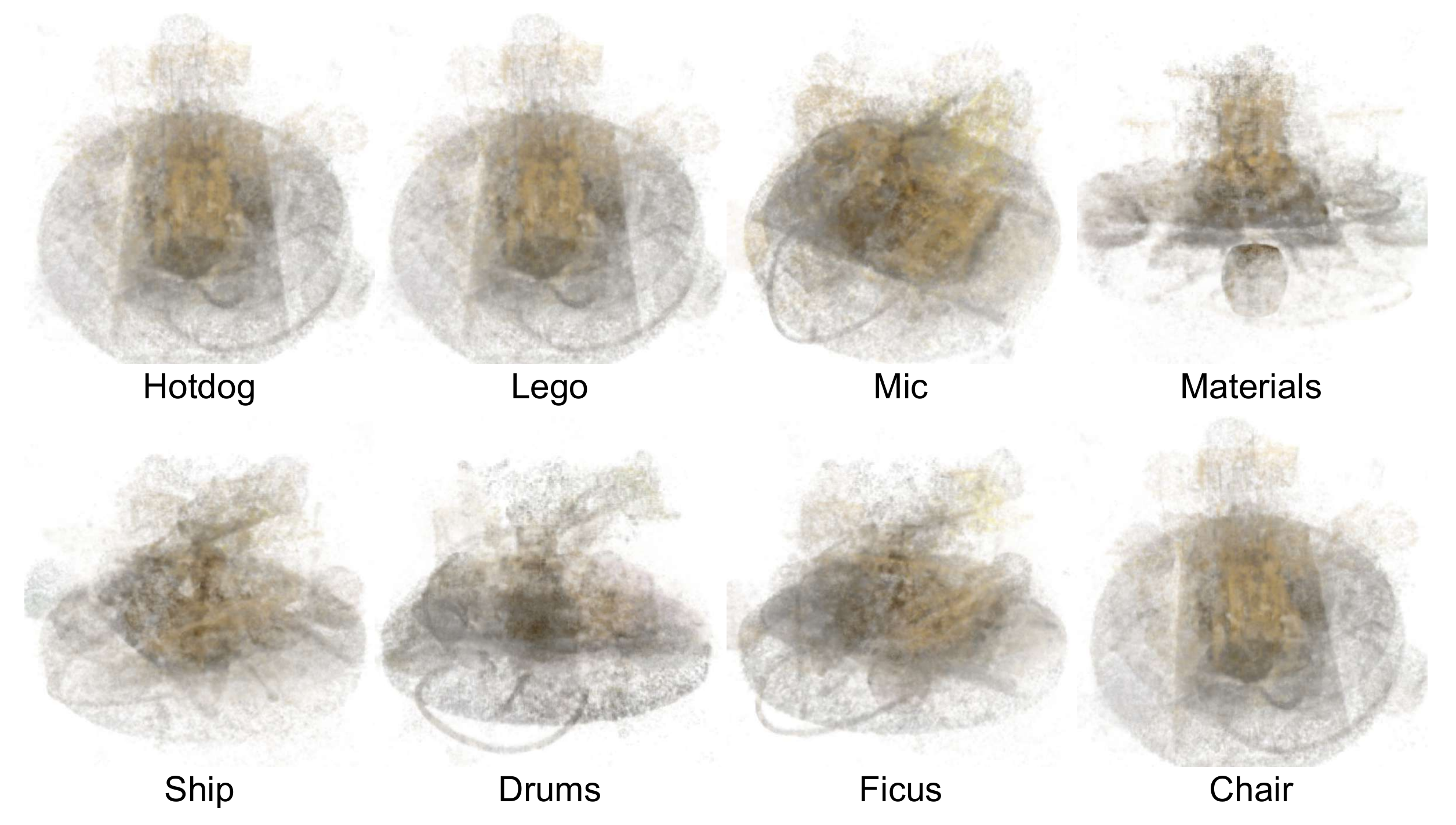}   &  \includegraphics[width=0.48\linewidth]{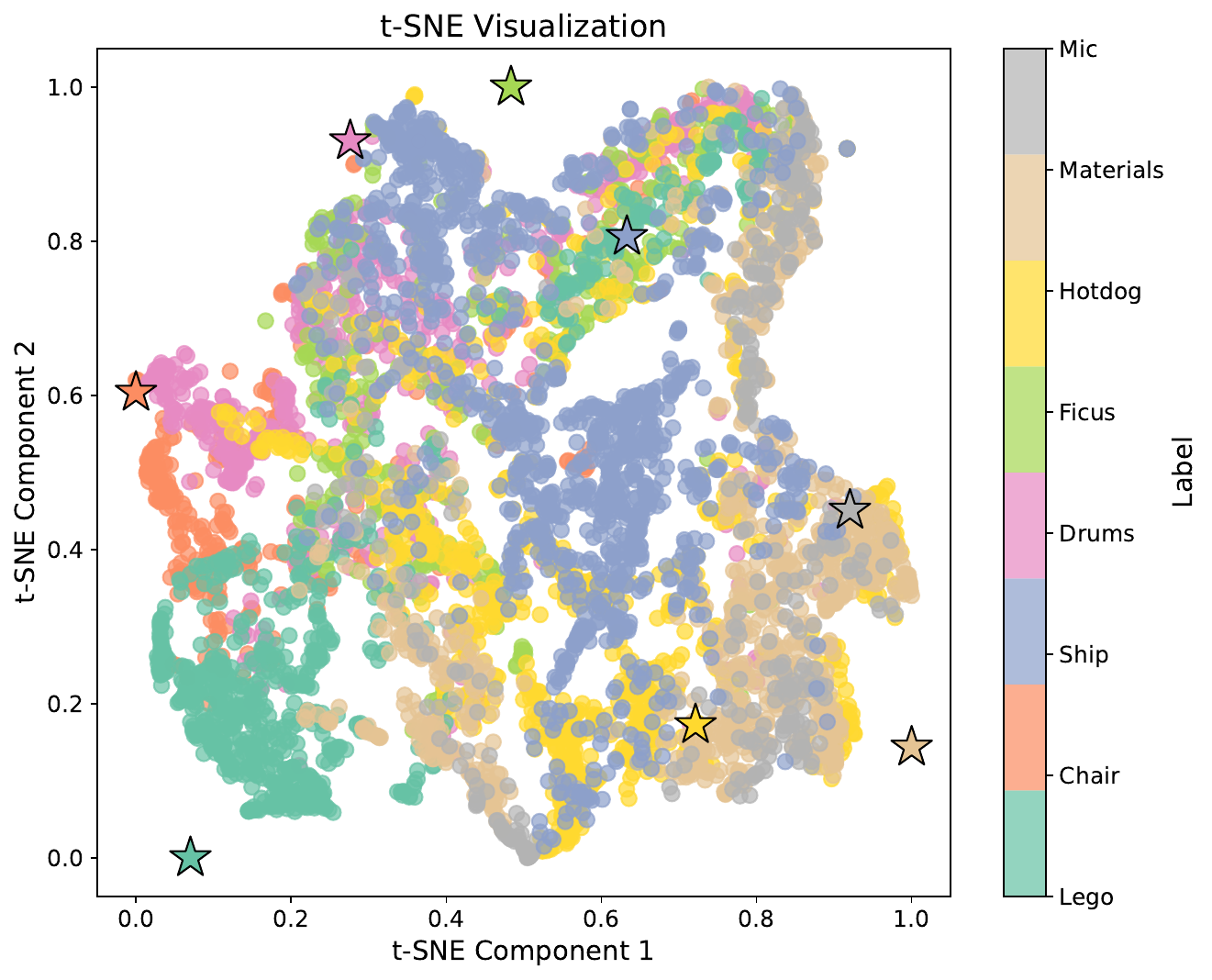}
            \end{tabular}
        }
        \vspace{0.5em}
        \captionof{figure}{(Left) We combine different scenes of the NeRF Synthetic $360^{\circ}$ \cite{mildenhall2021nerf} dataset and train Instant-NGP. The results show that neural hash encoding of scene coordinates may not be sufficient to preserve information across multiple scenes. (Right) The figure illustrates the representation space of C-NGP, visualized through a t-SNE map over scenes from NeRF Synthetic $360^{\circ}$ dataset \cite{mildenhall2021nerf}.}
        \label{fig:basengp_tsne}
    \end{minipage}
\end{figure}

\section{Shared Representation Space}
\label{sec:sharedspace}

As shown in Figure \ref{fig:basengp_tsne} (left), training Instant-NGP \cite{muller2022instant} on a mixed dataset without conditioning led to the intermixing of scene information. This confirms that neural hashing with only scene coordinates is insufficient for distinguishing scenes in a multi-scene setting. To achieve unique scene representations, additional conditioning is required.

Figure \ref{fig:basengp_tsne} (right) shows the t-SNE map \cite{van2008visualizing} of the learned multi-scene representation space of C-NGP across eight scenes from the NeRF Synthetic $360^{\circ}$ dataset \cite{mildenhall2021nerf}. The overlapping scene clusters suggest that the parameter space is shared across multiple scenes, meaning a subset of parameters encodes shared information rather than being exclusive to a single scene.

To generate this visualization, we extract neural hash encodings from one image per scene, all captured from the same viewpoint. Since each image is $800 \times 800$ pixels and many pixels correspond to the background, we divide it into $50 \times 50$ patches, yielding 16 patches per image. Instead of extracting features from every pixel, we obtain MLP features from these patches across all eight scenes and visualize them using t-SNE \cite{van2008visualizing}.

\begin{figure}[!t]
\begin{minipage}{\linewidth}
\begingroup
\removelatexerror
\begin{algorithm}[H]
\DontPrintSemicolon  
  \KwInput{$I_{new}$: list of new scene images}
  \KwInput{$l_{new}$: pseudo label for new scene}
  \KwInput{$L_{prev}$: list of pseudo labels for previous scenes}
  \KwInput{$F_{\theta}$: NeRF network}
  \KwInput{$C_{I_{new}}$: camera parameters of new scene} 
  \KwInput{$k$: number of images to render for previous scenes} 
  \tcc{Render previous scenes and store in $I_{prev}$}
  Initialize $I_{prev} \xleftarrow{} \{\}$ \\
  \For{$l^{i}_{prev} \in L_{prev}$}{
    $render\_image \xleftarrow{} F_{\theta}(l^{i}_{prev}, C_{I_{new}}, k)$ \\
    $I_{prev}$.append($render\_image$) \\
  }
  $T_{images} \xleftarrow{} concat(I_{new}, I_{prev})$ \tcp*{train images list}
  $T_{labels} \xleftarrow{} concat(L_{new}, L_{prev})$ \tcp*{train labels list}
  \tcc{$TrainNeRF()$ train the $F_{\theta}$ on given scene list and labels}
  $F_{\theta}^{updated} \xleftarrow{} TrainNeRF(F_{\theta}, T_{images}, T_{labels}, C_{I_{new}})$ \\
  \KwOutput{Updated $F_{\theta}^{updated}$ on new scene}
\caption{Learning new scenes using Conditional-cum-Continual NeRF (C-NGP)}
\label{algo:c3nerf}
\end{algorithm}
\endgroup
\end{minipage}
\end{figure}

\section{Architecture Choice}
\label{sec:ablation}

We conducted an ablation study to determine the most effective way to integrate multi-scene conditioning into the Instant-NGP \cite{muller2022instant} architecture, and the results are summarized in the main paper Table $3$ (right). 

The baseline, \underline{XYZ | C}, feeds the pseudo label alongside $(x, y, z)$ coordinates into the neural hashing module. While straightforward, this approach relies solely on raw labels for scene differentiation. In \underline{XYZ + $\psi(C)$}, we replaced the raw label with its positional encoding $\psi(C)$, but this led to lower performance, indicating that direct pseudo-label input is crucial for effective scene discrimination.

Adding both the raw label and its positional encoding (\underline{XYZ | C + $\psi(C)$}) improved results, suggesting that combining these representations enhances scene modeling. The best performance was achieved with \underline{XYZ | C + $\psi(C)$ + $SE(\theta, \phi)$ + $\psi(C)$}, where we further incorporated the spherical encoding of the viewing direction $SE(\theta, \phi)$ along with an additional instance of $\psi(C)$. This setup yielded the highest PSNR and SSIM scores and the lowest LPIPS, highlighting the benefit of jointly leveraging scene coordinates, viewing directions, and both raw and encoded labels.

Overall, our study shows that enriching multi-scene representations with positional and spherical encodings significantly improves scene modeling and retention.

\begin{figure*}[!t]
  \centering
  \includegraphics[width=1.0\linewidth]{images/nerf_synth.pdf}\vspace{0.7em}
  \caption{Qualitative demonstration of the quality of rendered images across different views through C-NGP over the scenes from the NeRF Synthetic $360^{\circ}$ dataset \cite{mildenhall2021nerf}. It is best viewed in PDF with Zoom.}
  \label{fig:nerf_synth_gallery}
\end{figure*}

\section{Additional Discussion}

Empirically, we explored the upper bound on the number of scenes that a fixed number of parameters can model. Although information forgetting can be minimized further with an increase in the number of parameters and scenes, our goal was to maximally utilize the representation capacity of a single neural radiance field parameter. Interestingly, we show that online sampling under generative replay produces better results compared to offline sampling (algorithm \ref{algo:c3nerf}). While the proposed framework can be viewed as taking the first steps toward multi-scene modeling, several open research avenues exist around it. An interesting future work could involve studying if the model learns useful scene priors during a continual learning process to aid in either faster learning over new scenes (\textit{i.e.}, convergence in fewer iterations or higher initial PSNR while learning to model a new scene) or can model new scenes with fewer number of images (few-shot learning). Moreover, dissecting C-NGP or the Instant-NGP backbone to analyze the kind of scene attributes handled by the model at each layer to increase physical interpretability is also an interesting avenue to explore. With the current exploration of the representative capacity of NeRFs and the aforementioned future directions, we believe that this work could serve as a primer for a new perspective on designing multi-scene continual neural radiance fields to continue accommodating new scenes without or with minimal loss of information of previously learned scenes. 

\begin{figure*}[!t]
  \centering
  \includegraphics[width=1.0\linewidth]{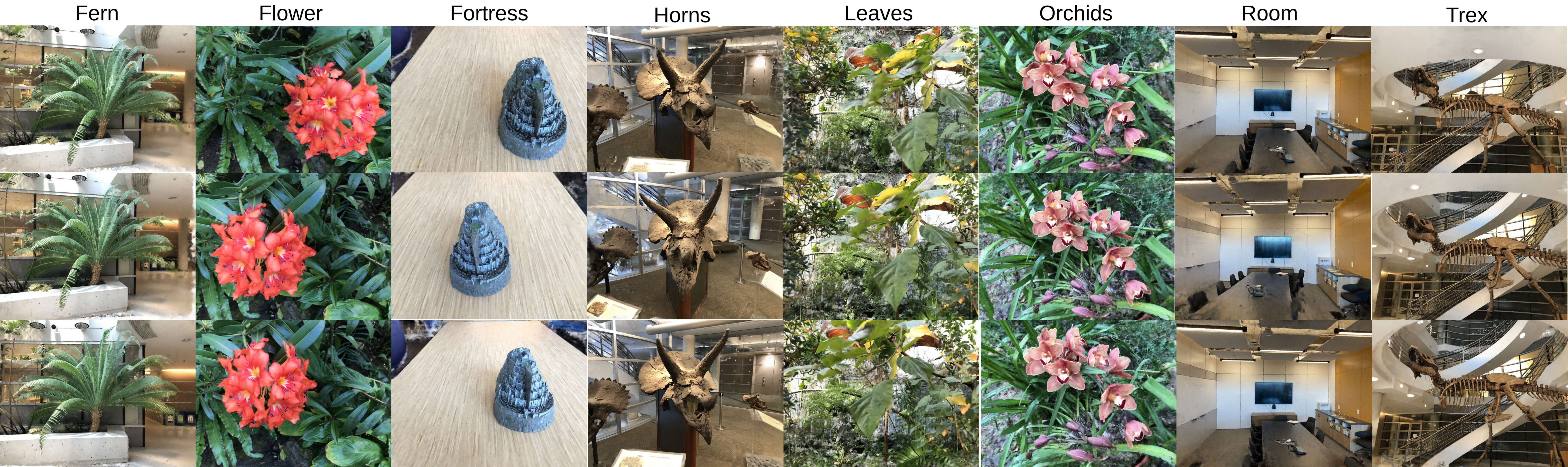}\vspace{0.7em}
  \caption{Qualitative demonstration of the quality of rendered images across different views through C-NGP over the scenes from the Real Forward-Facing dataset \cite{mildenhall2019local}. It is best viewed in PDF with Zoom.}
  \label{fig:nerf_real_gallery}
\end{figure*}

\begin{figure*}[!t]
  \centering
  \includegraphics[width=1.0\linewidth]{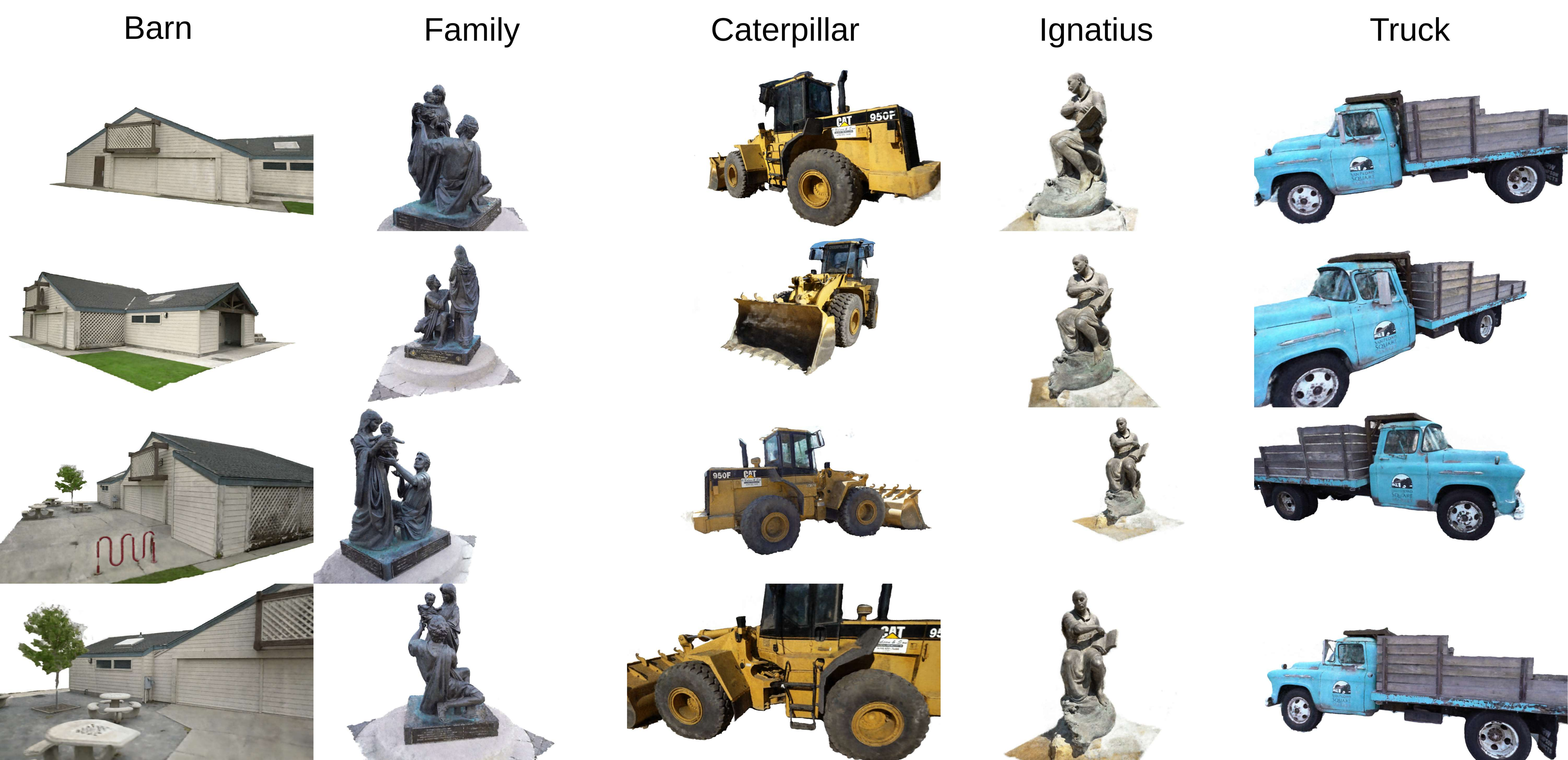}\vspace{0.7em}
  \caption{Qualitative demonstration of the quality of rendered images across different views through C-NGP over the scenes from the Tanks and Temples dataset \cite{knapitsch2017tanks}.}
  \label{fig:tanks_gallery}
\end{figure*}


\end{document}